\documentclass[iicol, oneside]{sn-jnl}

\usepackage{graphicx}%
\usepackage{multirow}%
\usepackage{amsmath,amssymb,amsfonts}%
\usepackage{amsthm}%
\usepackage{mathrsfs}%
\usepackage[title]{appendix}%
\usepackage{xcolor}%
\usepackage{textcomp}%
\usepackage{manyfoot}%
\usepackage{booktabs}%
\usepackage{algorithm}%
\usepackage{algorithmicx}%
\usepackage{algpseudocode}%
\usepackage{listings}%
\usepackage{hyperref}
\usepackage{cleveref}
\usepackage{colortbl}
\usepackage{hhline}
\usepackage{bm}
\usepackage{graphicx}  
\usepackage{subcaption} 
\definecolor{myGray}{gray}{0.9}
\makeatletter

\def\etal{\emph{et al. }}
\usepackage{pifont}  
\newcommand{\xmark}{\ding{55}}     

\AtEndPreamble{
    \crefname{section}{Sec.}{Secs.}
    \Crefname{section}{Section}{Sections}
    \crefname{table}{Tab.}{Tabs.}
    \Crefname{table}{Table}{Tables}
    \crefname{figure}{Fig.}{Figs.}
    \Crefname{figure}{Figure}{Figures}
}

\begin{document}

\title[Article Title]{Temporal Context Consistency Above All: Enhancing Long-Term Anticipation
by Learning and Enforcing Temporal Constraints}


\author*[1]{\fnm{Alberto} \sur{Maté}}\email{amate@iri.upc.edu}

\author[1]{\fnm{Mariella} \sur{Dimiccoli}}\email{mdimiccoli@iri.upc.edu}

\affil[1]{\orgdiv{Institut de Robótica i Informática Industrial}, \orgname{CSIC-UPC}, \orgaddress{\city{Barcelona}, \country{Spain}}}


\abstract{This paper proposes a method for long-term action anticipation (LTA), the task of predicting action labels and their duration in a video given the observation of an initial untrimmed video interval. We build on an encoder-decoder architecture with parallel decoding and make two key contributions. First, we introduce a bi-directional action context regularizer module on the top of the decoder that ensures temporal context coherence in temporally adjacent segments. Second, we learn from classified segments a transition matrix that models the probability of transitioning from one action to another and the sequence is optimized globally over the full prediction interval. In addition, we use a specialized encoder for the task of action segmentation to increase the quality of the predictions in the observation interval at inference time, leading to a better understanding of the past. We validate our methods on four benchmark datasets for LTA, the EpicKitchen-55, EGTEA+, 50Salads and Breakfast demonstrating superior or comparable performance to state-of-the-art methods, including probabilistic models and also those based on Large Language Models, that assume trimmed video as input. The code will be released upon acceptance.}




\maketitle

\vspace{-1.5em}
\section{Introduction}
\label{sec:intro}

For a real-world agent interacting with the environment or other agents/people in a dynamically changing world, LTA—the task of predicting the entire sequence of actions several minutes ahead given an initial untrimmed video interval—is crucial. Not surprisingly, since its introduction less than one decade ago \cite{farhaWhenWillYou2018}, the task is receiving increasing attention from the community. While early methods were based on separate steps for action segmentation in the observation interval and action anticipation in the rest of the video \cite{farhaWhenWillYou2018, keTimeConditionedActionAnticipation2019}, 
FUTR \cite{FUTR} proposed an integrated framework for action segmentation and anticipation consisting of an encoder-decoder transformer based architecture. 
Lately, increasing attention is being paid to the use of Large Language Models (LLM) to assist the task of LTA. Although these methods have proved to be effective for this goal, they assume trimmed video as input and typically do not predict duration, which is very important for LTA applications as planning.
\begin{figure}
  \centering
  \includegraphics[width=1\linewidth]{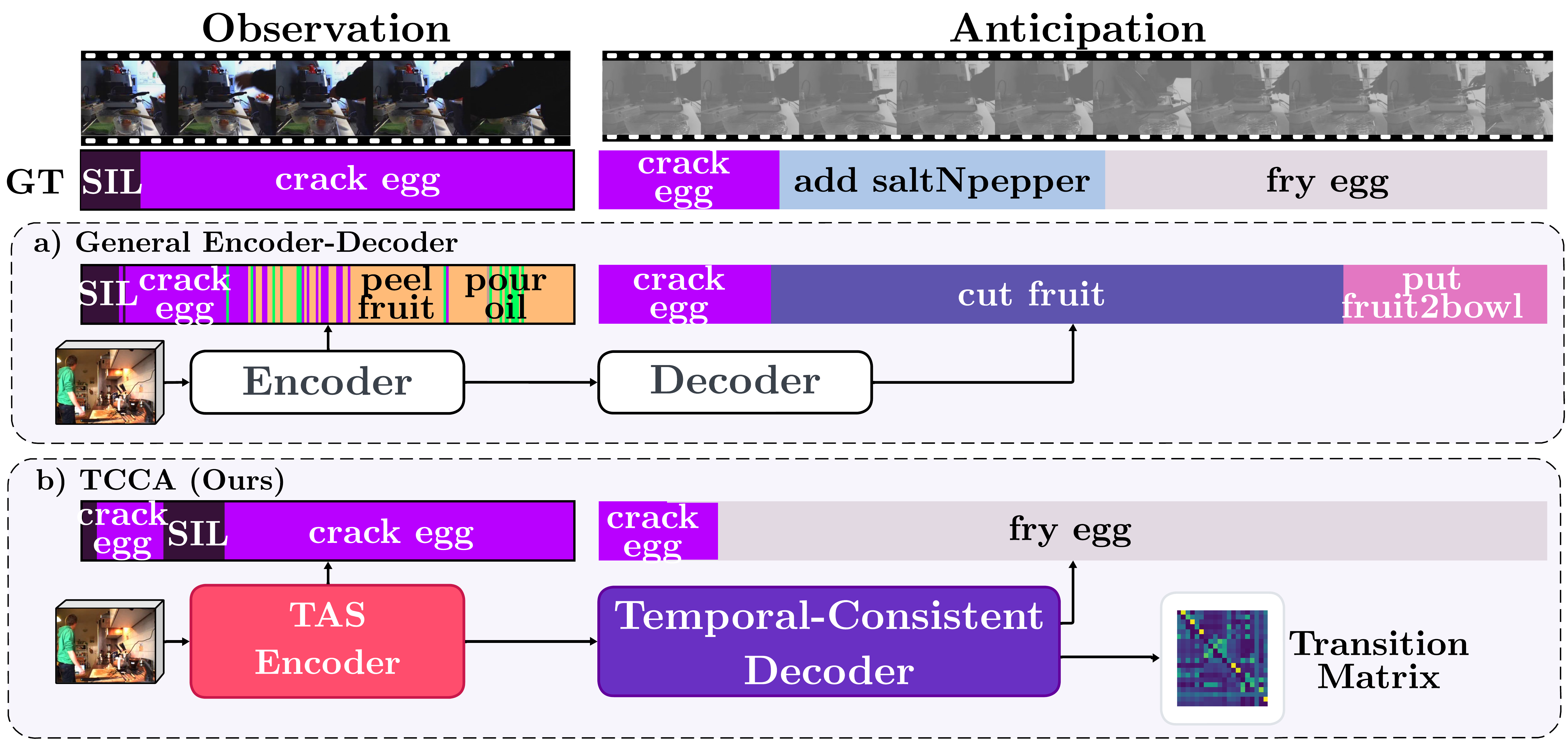}
    \caption{\textbf{Overview of our method}. Given the initial untrimmed portion of the video, LTA predicts future actions and their durations. Using the 'fried egg' activity from the Breakfast dataset \cite{breakfast}: (a) shows the results of \cite{FUTR}; (b) shows our method, TCCA, with an enhanced Temporal Action Segmentation (TAS) encoder and a temporal-consistent decoder, ensuring better results.}

   \label{fig:motivation}
   \vspace{-5mm}
\end{figure}

This paper builds on the joint framework for action segmentation and anticipation proposed in \cite{FUTR} and introduces two keys components aimed at keeping temporal context coherence of actions and their duration over the full anticipation interval. The main contributions are as follows:
\begin{itemize}
\item We introduce an end-to-end transformer based network, dubbed TCCA (Temporal Context Consistency above All), which effectively learns from the past untrimmed observations and leverages global and local interactions for long-term action anticipation.
\item We introduce a Bi-Directional Action Context Regularizer (BACR) to enforce local temporal context in parallel decoding. 
\item We propose to optimize globally the sequence of predicted actions by leveraging learned transition probabilities in parallel decoding.
\item The proposed method sets new state-of-the-art on four standard benchmarks for LTA: Breakfast, 50Salads, EpicKitchen-55, and EGTEA+ datasets.
\end{itemize}

\section{Related work}
\subsection{Long-term Action Anticipation (LTA)}
LTA seeks to predict sequences of future actions over extended periods, given an initial short portion of a video, that generally lasts a few minutes in total. The LTA task from untrimmed videos was introduced by Farha \etal \cite{farhaWhenWillYou2018} but it is rapidly gaining momentum due to its practical and scientific interest.  The seminal method \cite{farhaWhenWillYou2018} consists of a CNN-based model for action prediction in the observation interval and an RNN-based model for action anticipation that takes as input the labels generated by the CNN-based module. To account for uncertainty about the future, the same authors propose later to model the probability distribution of the future activities and to use this distribution to generate several possible sequences of future
activities at test time \cite{abu2019uncertainty}.  However, in both models, a predicted action segment sampled from the model and its duration are recursively combined with the observation to predict the next time step, resulting in accumulated prediction errors. To address this problem, Ke \etal \cite{keTimeConditionedActionAnticipation2019} developed a model that incorporates the time parameter to anticipate actions of any future time in one shot.
With the goal of demonstrating the relevance of both visual information and semantic labels for LTA, \cite{gammulle2019forecasting} proposed a two memory network architecture that takes as input the observed video sequence together with the observed label sequence (ground-truth) capturing different information cues to support the prediction task, and models long-term dependencies of individual streams through separate memory modules. 
While all these previous models separated the action segmentation task on the observed segment from the action anticipation in the future,  Farha \etal \cite{cycleConsistency} proposed an end-to-end model 
that directly maps the sequence of observed frames to a sequence of future activities and their duration, and subsequently predicts the past activities given the predicted future to check cycle consistency with the goal of detecting missed preceding actions.
Looking at the more general problem of untrimmed video understanding, Sener \etal \cite{senerTemporalAggregateRepresentations2020} proposed a multi-scale temporal aggregation method that models long-range interactions between recent and spanning representations pooled
from video snippets via coupled attention mechanisms. The authors also demonstrated the importance of correct action segmentation labels for good anticipation predictions. 
More recently, \cite{FUTR} proposed an encoder-decoder architecture based on transformers \cite{vaswani2017attention}, where the encoder captures fine-grained long-range temporal relations in the observation interval and the decoder learns to capture via cross-attention global relations between the observed features from the encoder and the upcoming actions.
Interestingly, this model introduced a query-based decoder that works in a parallel fashion to avoid accumulation errors which are typical of autoregressive models. 

Instead of tuning an encoder for action segmentation, the Anticipatr model \cite{ANTICIPATR} uses a two-stage learning approach, where the first stage trains a segment encoder to predict a set of future action instances, and the second stage uses it in conjunction with video-level encoder. Recently, \cite{gtda2024zatsarynna} introduced a Gated Temporal Diffusion (GTD) network that generates stochastic long-term anticipation by jointly modeling the uncertainty of both the observation and the future predictions through a generator backbone, which steers the information flow between observed and future frames. Interestingly, Gong et al. \cite{gong2024actfusion} proposed an unified diffusion model that jointly addresses TAS and LTA through a single training process and introduce anticipative
masking for effective task unification.

Lately, with the introduction of the Ego4D LTA challenge\footnote{https://eval.ai/web/challenges/challenge-page/1598/overview}, 
LLMs started to be exploited for LTA \cite{wang2023vamos,ANTGPT,zhang2024object}. The task is, given an input video up to a particular timestep (corresponding to the last visible action), to predict a list of $Z$ plausible action sequences. Among LLM models tailored to LTA, we highlight  ObjectPrompt \cite{zhang2024object}, AntGPT\cite{ANTGPT}, and PALM \cite{LALM}. 
ObjectPrompt \cite{zhang2024object}  learns, given object regions, task-specific object-centric representations from general-purpose pre-trained large vision-language models and dynamically associates motion with object evidence in order to predict future actions at different time steps through a predictive transformer encoder.
AntGPT\cite{ANTGPT} fine-tunes an LLM with an action recognition model to infer future actions from trimmed videos and demonstrates the usefulness of LLM to infer rare actions. PALM \cite{LALM} describes the past by using a hybrid vision-language model
in conjunction with an action recognition model, 
 and proposes a new strategy for in-context learning \cite{brown2020language}.

While LLM based methods have demonstrated to be largely useful in assisting LTA, they currently assume trimmed video as input and lack of a mechanism to automatically select the best possible sequence automatically. In addition, the estimated duration of future actions, when available, is not learned directly from visual content. 
Our proposed approach aims at addressing these limitations by optimizing globally the sequence of predicted actions and by learning the duration of future actions from input untrimmed videos. 

\begin{figure*}
  \centering
  \includegraphics[width=1\linewidth]{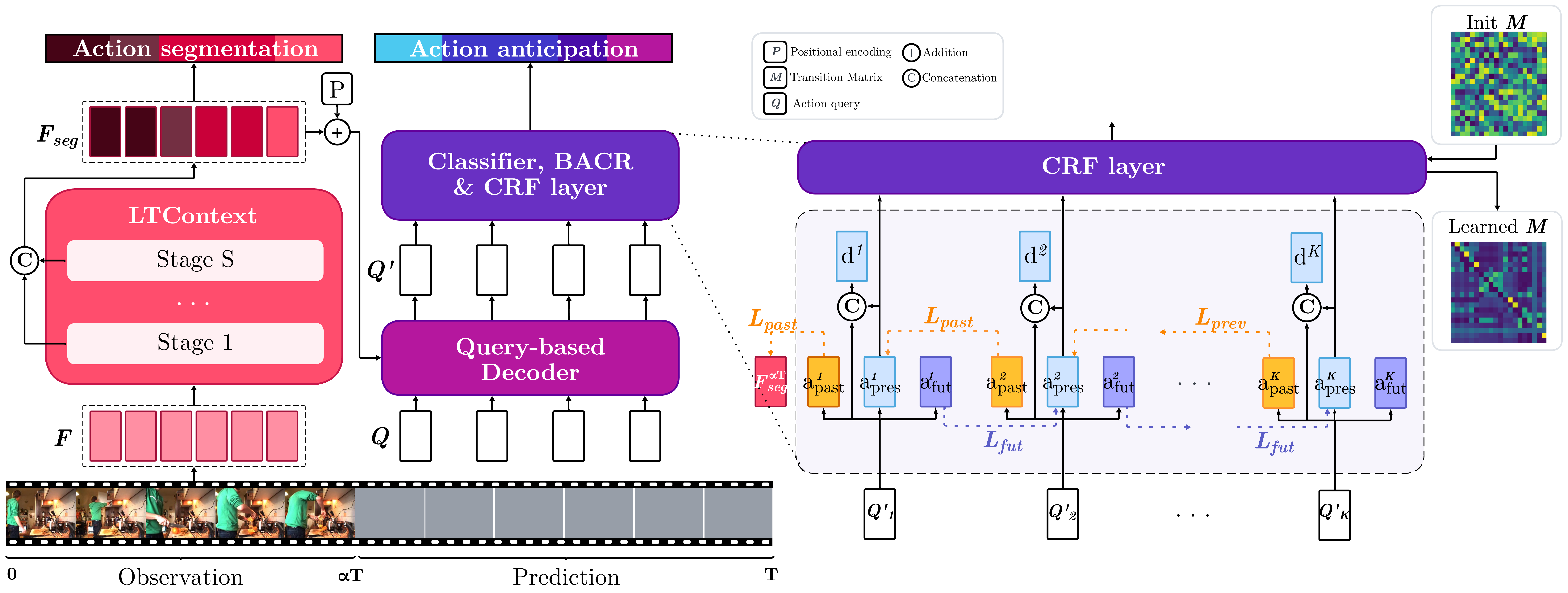}

    \caption{\textbf{Diagram of the TCCA Architecture.} Our method builds on an encoder-decoder structure. The LTContext encoder with temporal smoothing loss processes the observed portion of the video, transforming initial features $\bm{F}$ into logits $\bm{F_{\text{seg}}}$ representing action segmentation logits across $\bm{S}$ stages. These logits are used in the query-based decoder with cross-attention to queries $\bm{Q}$. The queries $\bm{Q^\prime}$ are classified by the Classifier \& BACR module and further processed by the CRF Layer to predict the future action sequence, utilizing a transition matrix $\bm{M}$ that is learnt through training. The right side details the Classifier, BACR and CRF, where $\bm{Q'}$ is projected into current, previous ($\bm{a_{past}^\prime}$), and next ($\bm{a_{fut}^i}$) action logits. These last two are trained with a KL loss between previous ($\bm{a_{pres}^{i-1}}$) and next ($\bm{a_{pres}^{i+1}}$) action logits, respectively. The duration of the $i-th$ segment is partially dependent on $\bm{a^i_{pres}}$ and $Q_i^\prime$.}
   \label{fig:method}
   \vspace{-4mm}

\end{figure*}

\subsection{Long-range Video Understanding}
Long-range video understanding is a problem that encompasses several computer vision tasks as action segmentation, action localization and action anticipation. The challenge of capturing long-range temporal dependencies was initially addressed by employing sequential models like hidden Markov model \cite{kuehne2016end}, Conditional Random Fields (CRF) \cite{mavroudi2018end} and recurrent networks \cite{kuehne2018hybrid}. However, these approaches experience forgetting issues with long videos and are computationally expensive. Deep learning based temporal models as Temporal Convolutional Network (TCN) \cite{lea2017temporal},  Multi-Stage TCN \cite{li2020ms}, and more recently Transformers \cite{UVAST, FACT} and Sparse Transformers \cite{LTContext} have been subsequently proposed. 
However, besides remarkable results, these models are still unable to learn logical temporal constraints from the data.
Xu \etal \cite{xu2022don} demonstrates that declarative temporal constraints can be extracted offline and explicitly provided to a deep network during training to reduce logical errors in its output. A shortcoming of this approach is that it requires a tedious curation of the dataset to encode the temporal logic constraints. Gong \etal 
\cite{gong2024activity} introduces an activity grammar induction algorithm that 
extracts a powerful probabilistic context-free grammar while capturing the characteristics of the activity.
The grammar enables recursive rules with flexible temporal orders, leading to powerful generalization capability.
Another approach proposed recently is to supplement an explicit representation of the statistical regularities in the target domain at
training time through a loss term that relies on an event transition matrix computed offline \cite{ETM}. Lately, LLMs are also being exploited to predict the goal from the observed part of the video and make goal coherent predictions \cite{mittal2024can}.

Instead, inspired by applications in NLP for Named Entity Recognition \cite{ huang2015bidirectionallstmcrfmodelssequence}, we propose a method that learns automatically from decoded segments a probability transition matrix of future actions by using a CRF model. We use it to optimize globally the sequence of future predictions.
Note that in \cite{mavroudi2018end}, mentioned above, a CRF model is proposed for the action segmentation task, by conditioning the probability of
action labels to the input features of frames (frame-level) and modeling them as a Gibbs distribution. Instead, we use  CRF for the first time for the LTA task, by conditioning action probabilities on
encoded tokens at the segment level, hence considerably reducing the computational cost.

\section{Methodology}

\paragraph{Problem definition.} The task we address is to predict future actions from a given observable portion of an untrimmed video sequence. Formally, given a video $V$ of length $T$, whose first $\alpha T$ frames are observable, with $\alpha \in [0,1]$, the objective is to forecast the sequence of $N$ actions for the subsequent $\beta T$ frames, with $\beta \in [0, 1-\alpha]$. $\alpha$ and $\beta$ are denoted as the observation ratio and the prediction ratio, respectively. 
The input consists of the observed video feature frames $\bm{F} \in \mathbb{R}^{\alpha T \times D}$, where $D$ is the dimension of the features, while the output is the predicted sequence of future $N$ action labels $\bm{a} \in \mathbb{R}^{N\times C}$, where $C$ is the total number of action classes, as well as their durations $\bm{d} \in \mathbb{R}^N$.

\subsection{Model}
We propose TCCA, the architecture depicted in Fig. \ref{fig:method}.
\vspace{-3mm}
\paragraph{Encoder.}
To effectively learn from the observation,  we used the LTContext encoder \cite{LTContext}, originally proposed for the task of action segmentation,  that leverages sparse attention mechanisms for capturing the full context of a video and windowed attention for capturing local dependencies.  Additionally, it employs a hierarchical approach with $\bm{S}$ stages for a refined action segmentation. The segmentation loss used, $\mathcal{L}_{seg}$, is a multi-stage framewise cross-entropy loss. This multi-stage strategy optimizes at various granularities, leveraging hierarchical context. 
Furthermore, to ensure that consecutive frames exhibit similar probability distributions and reduce over-segmentation, we integrate a temporal smoothness loss, $\mathcal{L}_{s}$, as introduced in \cite{MSTCN}. 

The encoder extracts action segmentation logits for each stage and creates $\bm{F_{seg}} \in \mathbb{R}^{\alpha T \times {D_{seg}}}$, where ${D_{seg}}=S\times C$, that captures past temporal action distributions to fed the decoder.

\paragraph{Decoder.}
\label{sec:method_decoder}

TCCA builds upon the parallel decoder architecture recently introduced in \cite{FUTR} and subsequently adopted by \cite{ANTICIPATR}, which predicts all actions in parallel during inference rather than auto-regressively like in \cite{UVAST, AVT}, hence avoiding accumulation errors. This architecture leverages a query-based decoder, originally used in object detection \cite{DETR} and includes learnable queries $\bm{Q} \in \mathbb{R}^{K \times D_{dec}}$, each representing an action segment to be predicted, where $K$ is the number of action queries. The number of actions $N$  estimated in each video is $N\leq{K}$ since after the \texttt{<EOS>} token, the remaining actions are discarded. The query based decoder also consists of multi-head cross-attention between encoder and decoder, along with self-attention mechanisms.
 The output features of the encoder ($\bm{F_{seg}}$) are first concatenated to a learned positional encoding, $\bm{P}$ and subsequently linearly projected to match the dimension of the decoder. 
\begin{equation}
    F^\prime_{seg} = \phi_{enc2dec}(F_{seg} + P)
     \label{eq:enc2dec}
\end{equation}

where $\phi_{enc2dec}: \mathbb{R}^{D_{seg}} \rightarrow \mathbb{R}^{D_{dec}}$. The resulting features $\bm{F^\prime_{seg}} \in \mathbb{R}^{\alpha T \times D_{dec}}$ are subsequently passed through the decoder using cross-attention with the queries $\bm{Q}$. The processed queries, denoted as $\bm{Q^\prime}$, for each action segment are utilized to predict the present action logits $\hat{\bm{a}}_{pres}$, and the duration $\bm{\hat{d}}$ through the following linear projections:
\begin{equation}
\hat{{a}}_{pres} = \phi_a(Q^\prime)
\end{equation}
\begin{equation}
\hat{d} = \phi_d([Q^\prime, \text{softmax}(\hat{{a}}_{pres})])
\end{equation}
where $[\cdot, \cdot]$ means vector concatenation. This method ensures that the duration prediction is partially dependent on the predicted action, which differs from employing two independent head layers as suggested in previous works such as \cite{FUTR}. 
The duration vector $\bm{\hat{d}} \in \mathbb{R}^{N}$ is first transformed exponentially and then normalized to ensure that all values sum to 1 \cite{FUTR, DIFFANT, farhaWhenWillYou2018}. The duration loss is calculated using the mean squared error (MSE):
\begin{equation}
\mathcal{L}_{dur} = \frac{1}{K} \sum_{i=1}^{K} (d_i - \hat{d_i})^2
\label{eq
}
\end{equation}
Here, $d_i$ and $\hat{d}_i$  represent the actual duration and the predicted duration
for the $i$-th segment. 

\paragraph{Bi-Directional Action Context Regularizer.}
\label{sec:BACR}
 This regularizer enhances the model's predictive capability by incorporating the local context of both preceding and succeeding actions for each action segment. This bi-directional approach aims to provide a richer contextual understanding. 

Given the processed queries $\bm{Q^\prime}$ for each action segment, we predict the next action label $\bm{a_{fut}}$ and the previous action label $\bm{a_{past}}$ using additional linear projection heads. 
The BACR loss functions aim to ensure that the predicted next action label $\bm{a_{fut}^i}$ for the $i$-th action segment aligns with the actual action label of the $(i+1)$-th segment, $\bm{a_{pres}^{i+1}}$. Similarly, the predicted previous action label $\bm{a_{past}^i}$ for the $i$-th action segment is compared with the actual action label of the $(i-1)$-th segment, $\bm{a_{pres}^{i-1}}$. The predictions are made as follows:
\begin{equation}
a_{fut} = \phi_{a_{fut}}(Q^\prime)
\end{equation}
\begin{equation}
a_{past} = \phi_{a_{past}}(Q^\prime)
\end{equation}

This alignment is supervised using the Kullback-Leibler (KL) \cite{kullback1951information} divergence loss.

\begin{equation}
\mathcal{L}_{fut} = \sum_{i=1}^{K-1} \text{KL}({a_{fut}^i} \parallel \hat{a}_{pres}^{i+1})
\label{eq:loss_next}
\end{equation}
\begin{equation}
\mathcal{L}_{past} = \text{KL}({a_{past}^{i=1}} \parallel {F_{seg}^{\alpha T}}) + \sum_{i=2}^{K} \text{KL}({a}_{past}^i \parallel \hat{a}_{pres}^{i-1})
\label{eq:loss_prev}
\end{equation}

To ensure continuity and context consistency, the first query in the previous head is not omitted (see first term from \cref{eq:loss_prev}). Instead, it is supervised using the last logits of the action segmentation encoder. For a multi-stage encoder like LTContext, we utilize the logits from the final stage. This approach helps the model maintain consistency with the last observable action, as in most cases, the last action from the observation continues into the first action query.

\paragraph{Global Temporal Sequence Optimization.}
To account for transition probabilities between actions, we incorporate a Conditional Random Field (CRF) \cite{CRF} layer in the decoder, that uses a learnable transition matrix $\bm{M}$ to model the probabilities of transitioning from one action to another. This enhances the temporal coherence of predicted action sequences considering the dependencies between successive actions.
Given an input $\bm{\hat{a}}_{pres} \in \mathbb{R}^{K \times C}$ that represents the emissions logits and a ground-truth sequence  $\bm{a} = (a^1, a^2, \ldots, a^N) \in \mathbb{R}^{N}$, the score is defined as:
\begin{equation}
s(\bm{\hat{a}}_{pres},\bm{a}) = \sum_{i=1}^{N} \bm{\hat{a}}_{pres}^{i,a^i} + \omega \sum_{i=0}^{N} \bm{M}^{a^i, a^{i+1}}
\label{eq:crf_score}
\end{equation}
where $\bm{M}$ is the transition matrix and $\omega$ is weights the importance of the transitions. The main advantage of the CRF layer is that it optimizes globally the sequence of predicted actions, resulting in contextually consistent action sequences. 
The training objective is to minimize the negative log-probability of the correct action sequence:
\begin{equation}
\begin{split}
\mathcal{L}_{crf} &= -\log(p(\bm{a}|\bm{\hat{a}}_{pres}))= \\
&= \log\left(\sum_{\bm{a^\prime} \in Y_{\bm{\hat{a}}_{pres}}} e^{s(\bm{\hat{a}}_{pres},\bm{a^\prime})}\right) - s(\bm{\hat{a}}_{pres},\bm{a}) 
\end{split}
\end{equation}
where $Y_{\bm{\hat{a}}_{pres}}$ represents all possible sequences for a given emission logits $\bm{\hat{a}}_{pres}$. This encourages the model to produce valid sequences of actions. During inference, the optimal action sequence $\bm{a}^*$ is found by maximizing the score by a Viterbi-like algorithm using dynamic programming \cite{viterbi}:
\begin{equation}
\bm{a}^* = \underset{\tilde{\bm{a}} \in Y_{\bm{\hat{a}}_{pres}}}{\arg\max} \; s(\bm{\hat{a}}_{pres}, \tilde{\bm{a}})
\end{equation}

\paragraph{Training Objective.}

The final loss function is defined as:
\begin{equation}
\mathcal{L} = \mathcal{L}_{seg} + \lambda \mathcal{L}_{s} + \mathcal{L}_{dur} +  \mathcal{L}_{fut} + \mathcal{L}_{past}+  \mathcal{L}_{crf} 
\label{eq:loss_total}
\end{equation}

where $\lambda$ controls the contribution of the smoothness term and is fixed to $0.2$.

\section{Experiments}
\subsection{Experimental Setup}
\label{sec:experiments_setup}
\begin{table*}[ht!]
    \setlength\tabcolsep{3pt}
    \resizebox{\linewidth}{!}{
    \begin{tabular}{@{}cclcccccccccccccccc@{}}
    \toprule
    \multirow{2}{*}{\rotatebox{90}{Input}}& \multirow{2}{*}{\shortstack{Type}} & \multirow{2}{*}{Methods} & \multicolumn{4}{c}{Breakfast $\beta$ ($\alpha$ = 0.2)} & \multicolumn{4}{c}{Breakfast $\beta$ ($\alpha$ = 0.3)} & \multicolumn{4}{c}{50Salads $\beta$ ($\alpha$ = 0.2)} & \multicolumn{4}{c}{50Salads $\beta$ ($\alpha$ = 0.3)} \\
    \cmidrule(lr){4-7} \cmidrule(lr){8-11} \cmidrule(lr){12-15} \cmidrule(lr){16-19}
    & & & 0.1 & 0.2 & 0.3 & 0.5 & 0.1 & 0.2 & 0.3 & 0.5 & 0.1 & 0.2 & 0.3 & 0.5 & 0.1 & 0.2 & 0.3 & 0.5\\
    \midrule

    \multirow{4}{*}{\rotatebox{90}{Pred. label}} & \multirow{4}{*}{Fisher} & RNN~\cite{farhaWhenWillYou2018} & 18.11 & 17.20 & 15.94 & 15.81 & 21.64 & 20.02 & 19.73 & 19.21 & 30.06 & 25.43 & 18.74 & 13.49 & 30.77 & 17.19 & 14.79 & 9.77 \\
    & & CNN~\cite{farhaWhenWillYou2018}  & 17.90 & 16.35 & 15.37 & 14.54 & 22.44 & 20.12 & 19.69 & 18.76 & 21.24 & 19.03 & 15.98 & 9.87 & 29.14 & 20.14 & 17.46 & 10.86 \\
    & & UAAA~\cite{abu2019uncertainty} & 16.71 & 15.40 & 14.47 & 14.20 & 20.73 & 18.27 & 18.42 & 16.86 & 24.86 & 22.37 & 19.88 & 12.82 & 29.10 & 20.50 & 15.28 & 12.31 \\
    & & Timecond.~\cite{keTimeConditionedActionAnticipation2019}  & 18.41 & 17.21 & 16.42 & 15.84 & 22.75 & 20.44 & 19.64 & 19.75 & 32.51 & 27.61 & 21.26 & 15.99 & 35.12 & 27.05 & 22.05 & 15.59 \\
    \midrule
    \multirow{4}{*}{\rotatebox{90}{Misc}} & GT label & TempAgg~\cite{senerTemporalAggregateRepresentations2020} & \textcolor{gray}{65.50} & \textcolor{gray}{55.50} & \textcolor{gray}{46.80} & \textcolor{gray}{40.10} & \textcolor{gray}{67.40} & \textcolor{gray}{56.10} & \textcolor{gray}{47.40} & \textcolor{gray}{41.50} & \textcolor{gray}{47.20} & \textcolor{gray}{34.60} & \textcolor{gray}{30.50} & \textcolor{gray}{19.10} & \textcolor{gray}{44.80} & \textcolor{gray}{32.70} & \textcolor{gray}{23.50} & \textcolor{gray}{15.30}\\

    & GT label & Zhao \etal~\cite{zhao2020diverse} (avg.) & \textcolor{gray}{72.22} & \textcolor{gray}{62.40} & \textcolor{gray}{56.22} & \textcolor{gray}{45.95} & \textcolor{gray}{74.14} & \textcolor{gray}{71.32} & \textcolor{gray}{65.30} & \textcolor{gray}{52.38} & \textcolor{gray}{46.63} & \textcolor{gray}{35.62} & \textcolor{gray}{31.91} & \textcolor{gray}{21.37} & \textcolor{gray}{46.13} & \textcolor{gray}{36.37} & \textcolor{gray}{33.10} & \textcolor{gray}{19.45}\\
     & LLM & ObjectPrompt  ~\cite{zhang2024object} & -- & -- & -- & -- & -- & -- & -- & -- & \textcolor{gray}{37.40}  & \textcolor{gray}{28.90}  & \textcolor{gray}{24.20}  & \textcolor{gray}{18.10}  & \textcolor{gray}{28.00}  & \textcolor{gray}{24.00} & \textcolor{gray}{24.30} & \textcolor{gray}{19.30} \\
    \midrule

     \multirow{9}{*}{\rotatebox{90}{Features}} & \multirow{2}{*}{Fisher}  & CNN~\cite{farhaWhenWillYou2018} & 12.78 & 11.62 & 11.21 & 10.27 & 17.72 & 16.87 & 15.48 & 14.09 & -- & -- & -- & -- & -- & -- & -- & -- \\
    & & TempAgg~\cite{senerTemporalAggregateRepresentations2020} & 15.60 & 13.10 & 12.10 & 11.10 & 19.50 & 17.00 & 15.60 & 15.10 & 25.50 & 19.90 & 18.20 & 15.10 & 30.60 & 22.50 & 19.10 & 11.20 \\[.4mm]
\arrayrulecolor{lightgray}\hhline{*{1}{~}*{18}{-}}\arrayrulecolor{black} \rule{0pt}{2.4ex}
    & \multirow{5}{*}{I3D} & 
Anticipatr~\cite{ANTICIPATR} & \textcolor{gray}{37.40} & \textcolor{gray}{32.00}  & \textcolor{gray}{30.30} & \textcolor{gray}{28.60} & \textcolor{gray}{39.90} & \textcolor{gray}{35.70} & \textcolor{gray}{32.10} & \textcolor{gray}{29.40} & \textcolor{gray}{41.10} & \textcolor{gray}{35.00} & \textcolor{gray}{27.60} & \textcolor{gray}{27.30} & \textcolor{gray}{42.80} & \textcolor{gray}{42.30} & \textcolor{gray}{28.50} & \textcolor{gray}{23.60} \\

       &  & TempAgg~\cite{senerTemporalAggregateRepresentations2020} & 24.20 & 21.10 & 20.00 & 18.10 & 30.40 & 26.30 & 23.80 & 21.20 & -- & -- & -- & -- & -- & -- & -- & -- \\
     &  & Cycle Cons~\cite{cycleConsistency} & 25.88 & 23.42 & 22.42 & 21.54 & 29.66 & 27.37 & 25.58 & 25.20 & 34.76 & 28.41 & 21.82 & 15.25 & 34.39 & 23.70 & 18.95 & 15.89 \\
     &  & A-ACT~\cite{gupta2022act} & 26.70 & 24.30 & 23.20 & 21.70 & 30.80 & 28.30 & 26.10 & 25.80 & 35.40 & \underline{29.60} & 22.50 & 16.10 & 35.70 & 25.30 & 20.10 & 16.30 \\
     &  & FUTR~\cite{FUTR} & 27.70 & 24.55 & 22.83 & 22.04 & 32.27 & 29.88 & 27.49 & 25.87 & \underline{39.55} & 27.54 & 23.31 & 17.77 & 35.15 & 24.86 & \textbf{24.22} & 15.26 \\
     &  & GTAN (determ.)~\cite{gtda2024zatsarynna} & \underline{28.80} & \textbf{26.30} & \textbf{25.80} & \textbf{26.00} & 35.50 & \underline{32.90} & \underline{30.50} & \textbf{29.60} & 36.70 & 27.70 & \underline{23.80} & 17.40 & 32.20 & 24.90 & 17.40 & 14.90 \\
     &  & ActFusion~\cite{gong2024actfusion} & 28.25 & 25.52 & 24.66 & 23.25 & \underline{35.79} & 31.76 & 29.64 & \underline{28.78} & \underline{39.55} & 28.60 & 23.61 & \underline{19.90} & \textbf{42.80} & \underline{27.11} & \underline{23.48} & \textbf{22.00} \\
     \arrayrulecolor{lightgray}\hhline{*{2}{~}*{17}{-}}\arrayrulecolor{black} \rule{0pt}{2.4ex} &  & TCCA (Ours) & \textbf{28.83} & \underline{25.93} & \underline{25.19} & \underline{23.41} & \textbf{36.69} & \textbf{33.39} & \textbf{31.96} & 27.63 & \textbf{41.99} & \textbf{33.69} & \textbf{25.85} & \textbf{21.96} & \underline{37.12} & \textbf{28.31} & 22.85 & \underline{18.43} \\

    \bottomrule

    \end{tabular}}
    \caption{Comparisons on the Breakfast~\cite{breakfast} and 50Salads~\cite{50salads} datasets using MoC (\%). The highest accuracy is indicated in bold, and the second highest accuracy is underlined. Rows shaded in gray are not directly comparable.}
    \label{tab:breakfast_50s}
    \vspace{-5mm}

\end{table*}

\paragraph{Datasets.}
We evaluate our method on four standard action anticipation datasets. The \textit{Breakfast} dataset \cite{breakfast} consists of 1,712 videos featuring 52 individuals preparing breakfast in various kitchens. Each video is segmented into one of 10 activities related to breakfast preparation and annotated with 48 fine-grained action labels. Videos have an average duration of 2.3 minutes and a highly imbalanced action distribution \cite{TASsurvey}.
The \textit{50Salads} dataset \cite{50salads} comprises 50 top-down view videos of people preparing salads. It includes over 4 hours of RGB-D video data annotated with 17 fine-grained action labels. Videos are typically longer than Breakfast, averaging around 20 actions each.
 The \textit{EGTEA Gaze+} \cite{EGTEA} dataset contains 28 hours of video footage showcasing various actions with 10.3K action annotations. It includes 19 verbs, 51 nouns, and a total of 106 unique actions. Finally, the \textit{EpicKitchens-55} \cite{EGTEA} covers 55 hours of video, including 9,596 video segments labeled with 125 verbs, 352 nouns, and 2,513 action combinations. More datasets and additional experiments are provided in the supplementary material.

\paragraph{Metrics.}
Following previous work, we utilize different evaluation metrics specific to each dataset.
 For \textit{Breakfast} and \textit{50Salads}, we calculate the mean accuracy over classes (MoC), averaged frame-wise across all action classes for a given prediction horizon, as defined in \cite{farhaWhenWillYou2018}.
The initial observation window ($\alpha$) is set to either 20\% or 30\% of the video duration. The model then predicts actions for subsequent segments ($\beta$) of varying lengths (10\%, 20\%, 30\%, or 50\% of the entire video). The results are averaged over 4 splits for the Breakfast dataset and 5 splits for the 50Salads dataset.
For the \textit{EGTEA Gaze+} and \textit{EpicKitchens-55}  egocentric datasets, we use mean Average Precision (mAP), a multi-label classification metric following the protocol defined in \cite{EGOTOPO}. Here, a portion ($\alpha$) of each video serves as input, and the model predicts the set of all action classes for the remaining video segment (100\% - $\alpha$). We evaluate with $\alpha$ values of 25\%, 50\%, and 75\%, reporting mAP for all the target action (All) but also for low-shot (Rare) and many-shot (Freq) classes. Furthermore, following the previous protocol, the actions considered here are restricted to verb prediction. 
We also report the action segmentation from the past observation using their standard metrics: frame-wise accuracy (Acc), segmental Edit distance (Edit) and segmental F1 score (F1) at the overlapping thresholds of 10\%, 25\%, and 50\% denoted by F1@\{10,25,50\}.

\paragraph{Architecture Details.}
The overall architecture is shown in \cref{fig:method}. Our LTContext encoder follows the default configuration for Breakfast from \cite{LTContext} with some modifications: we changed the number of attention heads to 4 and the number of layers to 3 for all datasets. The hidden dimension is set to 64 for Breakfast and EpicKitchens-55, and 128 for 50salads and EGTEA. The encoder has 6 stages for 50salads and 4 stages for the other datasets.
The decoder is initialized with a hidden dimension that is twice the size of the encoder's hidden dimension. The number of decoder layers is 2 for Breakfast and EGTEA, and 3 for 50salads and EpicKitchens-55. The number of anticipation queries $\bm{K}$ is based on the number of actions to predict, initialized as 8, 20, 200, and 30 for Breakfast, 50salads, EpicKitchens-55, and EGTEA respectively. We use GeLU \cite{GELU} as the activation function for both encoder and decoder.

For the EGTEA and EpicKitchens-55 datasets, due to the nature of the task \cite{EGOTOPO}, we added a multi-label classification head for each action query to predict the set of future actions.
Additionally, we did not compute the duration. Both modifications were made because the mAP metric does not account for the order of the sequence since it was originally formulated for multi-label classification tasks and not for long-term prediction \cite{LALM}.

\paragraph{Training.}
We use the pre-extracted I3D \cite{I3D} as the visual input features $\bm{F}$ for all datasets, obtained from \cite{MSTCN, EGOTOPO}. The training process is made using an AdamW optimizer \cite{ADAMW} with a 1e-3 learning rate with a cosine annealing warm-up scheduler \cite{WARMPUP} with warm-up stages of 10 epochs. The number of epochs is 50, 35, 60, 50 for Breakfast, 50salads, EpicKitchens-55, and EGTEA respectively. We use a sample rate of 3 for Breakfast and 50salads, 2 for EpicKitchens-55 and 1 for EGTEA, selecting frames from a uniform distribution within these rates to reduce overfitting. The $\omega$ hyperparameter is set to 0.1 for Breakfast and 1 for the rest and the $\lambda$ is equal to 0.2 for all datasets. In the training stage, we use $\alpha \in \{20, 30, 50\}\%$  for Breakfast; $\alpha \in \{20, 30, 40, 50\}\%$ 50salads; $\alpha \in \{20-80\}\%$  in 10\% increments for EpicKitchens-55 and EGTEA.

\begin{table}[t]
    \centering
    \setlength\tabcolsep{5pt}
    \resizebox{1\linewidth}{!}{
    \begin{tabular}{@{}clcccccc@{}}
        \toprule
        \multirow{2}{*}{\rotatebox{90}{Type}}&\multirow{2}{*}{Method} & \multicolumn{3}{c}{EpicKitchens-55} & \multicolumn{3}{c}{EGTEA Gaze+} \\
        \cmidrule(lr){3-5} \cmidrule(lr){6-8}  
        & & All & Freq & Rare & All & Freq & Rare \\ \midrule
        \multirow{2}{*}{\rotatebox{90}{LLM}} & AntGPT~\cite{ANTGPT} & \textcolor{gray}{40.2} & \textcolor{gray}{58.8} & \textcolor{gray}{32.0} & \textcolor{gray}{80.2} & \textcolor{gray}{84.5} & \textcolor{gray}{74.0} \\
         & PALM~\cite{LALM} & \textcolor{gray}{40.4} & \textcolor{gray}{59.3} & \textcolor{gray}{30.3} & \textcolor{gray}{80.7} & \textcolor{gray}{85.0} & \textcolor{gray}{73.5} \\
        \midrule
        \multirow{7}{*}{\rotatebox{90}{I3D Features}} & I3D~\cite{I3D} & 32.7 & 53.3 & 23.0 & 72.1 & 79.3 & 53.3 \\
         & ActionVLAD~\cite{girdhar2017actionvlad} & 29.8 & 53.5 & 18.6 & 73.3 & 79.0 & 58.6 \\
         & Timeception~\cite{hussein2019timeception} & 35.6 & 55.9 & 26.1 & 74.1 & 79.7 & \underline{59.7} \\
         & VideoGraph~\cite{hussein2019videograph} & 22.5 & 49.4 & 14.0 & 67.7 & 77.1 & 47.2 \\
         & EGO-TOPO~\cite{EGOTOPO} & 38.0 & 56.9 & \underline{29.2} & 73.5 & {80.7} & 54.7 \\
         & Anticipatr~\cite{ANTICIPATR} & \underline{39.1} & \underline{58.1} & 29.1 & \underline{76.8} & \underline{83.3} & 55.1 \\

        \arrayrulecolor{lightgray}\hhline{*{1}{~}*{7}{-}}\arrayrulecolor{black} \rule{0pt}{2.4ex} & TCCA (Ours) & \textbf{40.6} & \textbf{59.4} & \textbf{30.7} & \textbf{80.3} & \textbf{85.8} & \textbf{65.7} \\
        \bottomrule
    \end{tabular}}
    \caption{Comparisons  on EpicKitchens-55 and EGTEA Gaze+ using mAP (\%) for the All, Rare and Freq action categories. The highest accuracy is indicated by bold numbers, and the second highest accuracy is underlined. Rows shaded in gray are not directly comparable because they use LLMs.}
    \label{tab:sota_ek55_egtea}
    \vspace{-6mm}
\end{table}
\subsection{Comparative Results}
\label{sec:results}

We compare TCCA with state-of-the-art approaches using the Breakfast and 50Salads benchmarks in \cref{tab:breakfast_50s}.  Methods using ground-truth labels as input \cite{senerTemporalAggregateRepresentations2020, zhao2020diverse} are listed as an upper-bound reference. These outstanding results show that improving the action segmentation could improve anticipation performance \cite{ANTSurvey}. Our method is directly compared to other methods that take as input I3D features of an untrimmed video interval. Note that Anticipatr \cite{ANTICIPATR} is not directly comparable because it uses a different evaluation protocol as it measures the predictions at $\beta$ of the residual video segments, as also noted in \cite{ANTSurvey, gtda2024zatsarynna}.  
Our approach achieves the best performance in 5 out of 8 metrics in 50salads dataset and has competitive results in the Breakfast dataset in particular when $\alpha=0.3$, outperformimg recent state-of-the-art methods based on diffusion models such as GTAN \cite{gtda2024zatsarynna} and ActFusion \cite{gong2024actfusion}.Compared with FUTR \cite{FUTR}, which uses a similar query-based decoder, the highest performance of TCCA demonstrate that our additional temporal understanding layers on top of the decoder enhance action anticipation by maintaining temporal consistency between action queries. Additionally, we observe that the results of FUTR \cite{FUTR} on50Salads reported here, are higher than those reported on the official GitHub repository \cite{ANTSurvey, gtda2024zatsarynna}. While not directly comparable, our method outperforms ObjectPrompt \cite{zhang2024object}, an LLM-based approach requiring trimmed video input.

We further evaluated our method on EpicKitchens-55 and EGTEA Gaze+ and results are presented in \cref{tab:sota_ek55_egtea}. Recent methods leveraging LLMs for LTA are included at the top of the table for a broader overview, despite differences in input and the high cost of fine-tuning these models (e.g., \cite{ANTGPT}). Our method outperforms current state-of-the-art approaches in all metrics for both datasets, even surpassing in some cases methods based on LLMs that assume trimmed videos as input instead of untrimmed ones. Even under these unbalanced conditions, for the Rare metric, our method achieves competitive results. 
\subsection{Ablation Analysis}
\label{sec:ablation}
We performed comprehensive analyses to validate the proposed method. In the subsequent experiments, we evaluated our method using the different datasets averaged over all the splits. The default settings are indicated in \colorbox{myGray}{gray}.
\vspace{-1.5mm}

\paragraph{Segmentation Loss.}
\label{p:segmentation_loss}
To evaluate the importance of understanding the past in the anticipation task, we summarize anticipation using the mMoC metric, calculated as the mean of all MoC values across the observation and prediction ratios. Additionally, we assess segmentation metrics within the observation horizon. Results from \cref{tab:ablation_seg} show that good action segmentation is really important for the task of anticipation. We also see that the $\mathcal{L}_{s}$ improves the segmentation of the past slightly, leading to less noisy features for the decoder and in turn incrementing the anticipation metric.

\begin{table}[t]
    \centering
    \setlength\tabcolsep{6pt}
    \resizebox{1\linewidth}{!}{
    \begin{tabular}{@{}cccccc@{}}
        \toprule
        \multicolumn{2}{c}{Loss}  & Ant. & \multicolumn{3}{c}{Segmentation} \\
        \cmidrule(lr){1-2} \cmidrule(lr){3-3} \cmidrule(lr){4-6} 
        $\mathcal{L}_\text{seg}$ & $\mathcal{L}_\text{s}$ & mMoC  & Acc & Edit & F1@$\{10, 25, 50\}$  \\
        \midrule
        \xmark & \xmark & 25.51 & - & - & -/-/- \\
        \checkmark & \xmark & \underline{27.95} & \underline{54.37} & \underline{60.57} & 63.92/55.53/41.70 \\
        \rowcolor{myGray} \checkmark & \checkmark & \textbf{29.13} & \textbf{54.53} & \textbf{62.01} & \textbf{64.26/55.69/41.86} \\
         \bottomrule
    \end{tabular}}
    \caption{Ablation of the segmentation losses on the Breakfast dataset. mMoC is the mean of MoC over observed and predicted horizons. We also report the segmentation metrics for the encoder.}
    \label{tab:ablation_seg}
    \vspace{-2mm}
\end{table}

\begin{table}[t]
    \centering
    \setlength\tabcolsep{3pt}
    \resizebox{1\linewidth}{!}{
    \begin{tabular}{@{}lccccc@{}}
        \toprule
        \multirow{2}{*}{Encoder}  & Ant. & \multicolumn{3}{c}{Segmentation} &  \multirow{2}{*}{Params.}  \\
        \cmidrule(lr){2-2} \cmidrule(lr){3-5} 
         &  mMoC  & Acc & Edit & F1@$\{10, 25, 50\}$ &   \\
        \midrule
        DETR \cite{DETR} & \underline{25.59} & 53.30 & 29.30 & 34.82/30.53/23.28 & 1.3M \\
        FACT \cite{FACT} & 23.40 & \underline{53.70} & \underline{51.37} & 55.79/49.32/37.86 & 1.3M \\
        \rowcolor{myGray} LTContext \cite{LTContext} & \textbf{29.13} & \textbf{54.53} & \textbf{62.01} & \textbf{64.26/55.69/41.86} & 1.3M \\
         \bottomrule
    \end{tabular}}
    \caption{Ablation of the encoder architecture for Breakfast. mMoC is the mean of MoC over observed and predicted horizons. We also report the segmentation metrics for the encoder.}
    \label{tab:ablation_encoder}
    \vspace{-5mm}
\end{table}

\begin{figure*}[t]
    \centering
    \begin{subfigure}[t]{\linewidth}
    \includegraphics[width=\columnwidth]{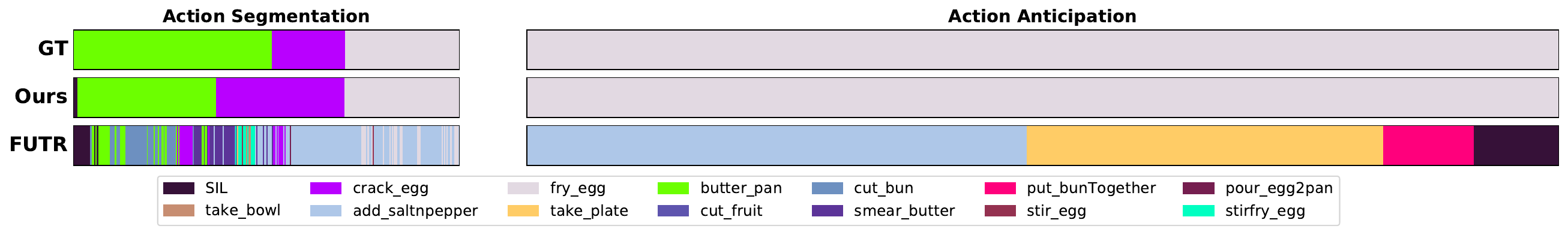}
    \caption{\textbf{Breakfast.} P49\_webcam02\_P49\_friedegg; ($\alpha$=0.2, $\beta$=0.5)}
    \label{fig:qua_a}
    \end{subfigure}

    \begin{subfigure}[t]{\linewidth}
    \includegraphics[width=\columnwidth]{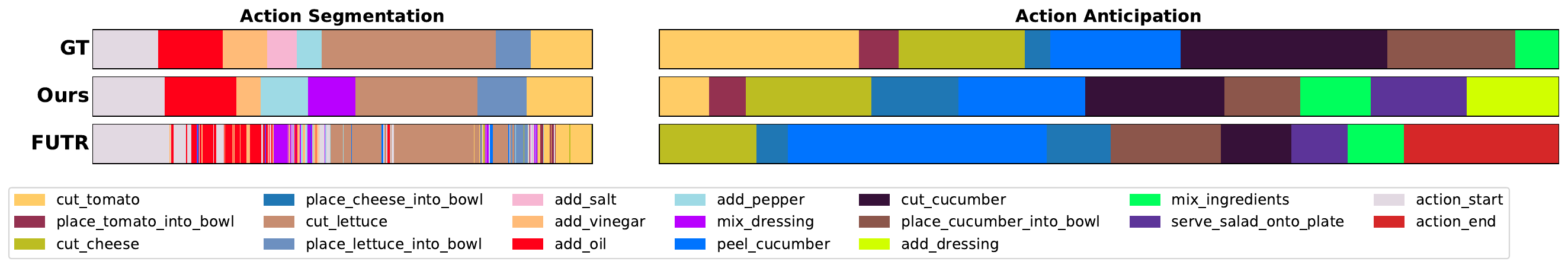}
    \caption{\textbf{50salads.} rgb-23-1; ($\alpha$=0.3, $\beta$=0.5)}
    \label{fig:qua_b}
    \end{subfigure}

    \vspace{-2mm}
\caption{\textbf{Qualitative results.} We display the ground-truth (GT), the results of TCCA (Ours) and FUTR \cite{FUTR} after using their official checkpoints. The left side of the diagram displays action segmentation from the observation, while the right side shows action anticipation after decoding the action and duration into a frame-wise sequence.
}
\vspace{-5mm}
 \label{fig:qualitative}
\end{figure*}

\paragraph{Encoder Architecture.}
 We experimented with multiple encoder architectures. We include the standard DETR \cite{DETR} encoder used in previous works \cite{FUTR, ANTICIPATR} and the recent TAS state-of-the-art architecture, FACT \cite{FACT}, that has the best segmentation metrics so far. This last encoder introduces a Frame-Action Cross-attention Temporal modeling framework, performing temporal modeling on both frame and action levels in parallel. This iterative bidirectional information transfer refines the features, enhancing accuracy. Results are shown in \cref{tab:ablation_encoder}. 

To ensure a fair comparison, we built all the encoders with a similar number of parameters. Although the frame-wise action segmentation results were similar across the three encoders, the segment-wise metrics (i.e., Edit and F1) indicated some over-segmentation, even when using $\mathcal{L}_{s}$. This was expected with DETR, as it was not explicitly designed for action segmentation. The issue with FACT may stem from the author's own claim that "the performance of FACT is more robust on longer videos" \cite{FACT}. Conversely, LTContext demonstrated in its original paper that it could achieve high F1@50 results even when the input window size was reduced by up to 12\% \cite{LTContext}. This results in segmentation lower the performance for the anticipation part. The superior performance of LTContext might be also due to its ability to use logits from different stages to build $\bm{F}_{seg}$, providing a hierarchy of results for the decoder, while the other methods rely solely on the final stage.

\begin{figure}
  \centering
  \includegraphics[width=1\linewidth]{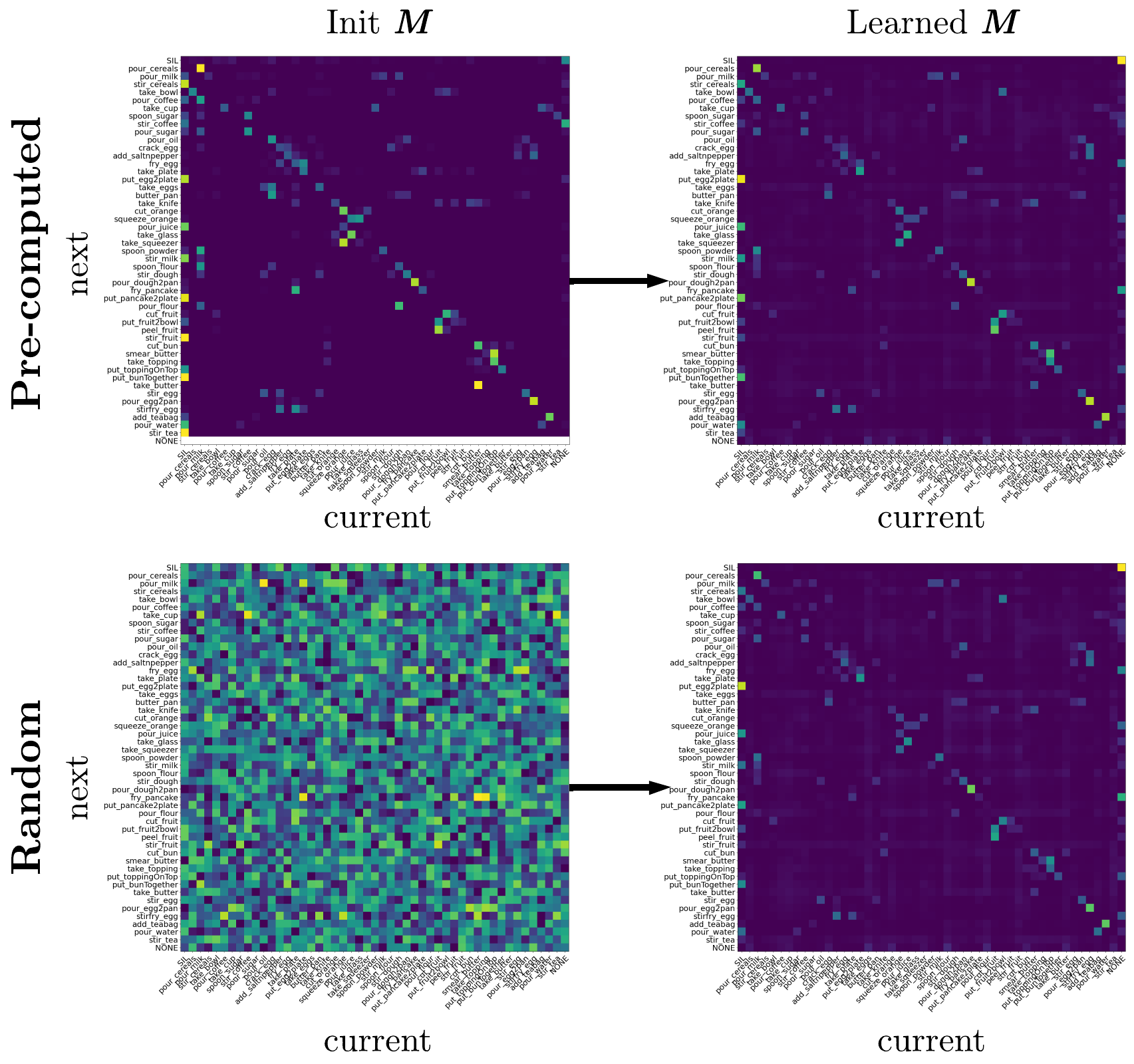}
    \caption{\textbf{Impact of transition matrix initialization.} We initialized $\bm{M}$ using a pre-computed matrix (top-left) based on statistical analysis, and a random matrix (bottom-left). The learned transition matrix from a pre-computed matrix (top-right) and random matrix (bottom-right) are remarkably similar.}
   \label{fig:matrix}
   \vspace{-7mm}
\end{figure}
\paragraph{BARC Loss.}
We first evaluated the impact of the BACR module over Breakfast in \cref{tab:ablation_BACR}. Combining both $\mathcal{L}_{prev}$ and $\mathcal{L}_{next}$ improves the results compared with just using one of them. This is because using both losses we let the queries be acknowledged about the future and the past. Additionally, we compared our proposed method BACR against the loss proposed in \cite{ETM}, where they use a transition matrix to compute the likelihood that each action is preceded or followed by every other action. \cref{tab:ablation_ETM} shows that our method outperforms ETM in all the prediction horizons.
\begin{table}[t]
    \centering
    \setlength\tabcolsep{8pt}
    \resizebox{0.85\linewidth}{!}{
    \begin{tabular}{@{}cccccc@{}}
        \toprule
        \multicolumn{2}{c}{Loss} & \multicolumn{4}{c}{$\beta$ ($\alpha$ = 0.3)}  \\
        \cmidrule(lr){1-2} \cmidrule(lr){3-6} 
        $\mathcal{L}_\text{prev}$ & $\mathcal{L}_\text{next}$  & 0.1 & 0.2 & 0.3 & 0.5  \\
        \midrule
        \xmark & \xmark & \underline{35.52} & 31.65 & 29.99 & 26.63 \\
        \xmark & \checkmark & 35.11 & 30.99 & 29.87 & 26.33 \\
        \checkmark & \xmark & 35.20 & \underline{31.96} & \underline{30.60} & \underline{26.86} \\
        \rowcolor{myGray} \checkmark & \checkmark & \textbf{36.69} & \textbf{33.39} & \textbf{31.96} & \textbf{27.63} \\
         \bottomrule
    \end{tabular}}
    \caption{Ablation of the BACR module on the Breakfast dataset.}
    \label{tab:ablation_BACR}
    \vspace{-4mm}
\end{table}

\begin{table}[t]
    \centering
    \setlength\tabcolsep{6pt}
    \resizebox{0.9\linewidth}{!}{
    \begin{tabular}{lcccc}
        \toprule
        \multirow{2}{*}{\shortstack{Method}} & \multicolumn{4}{c}{ $\beta$ ($\alpha$ = 0.3)} \\
        \cmidrule(lr){2-5} 
        & 0.1 & 0.2 & 0.3 & 0.5 \\
        \midrule
        ETM \cite{ETM} &  33.71 & 30.56 & 28.80 & 26.49  \\
        \rowcolor{myGray} BACR & \textbf{36.69} & \textbf{33.39} & \textbf{31.96} & \textbf{27.63} \\
         \bottomrule
    \end{tabular}}
    \caption{BACR versus the ETM loss on the Breakfast dateset.}
    \label{tab:ablation_ETM}
    \vspace{-4mm}

\end{table}
\begin{table}[t]
    \centering
    \setlength\tabcolsep{4pt}
    \resizebox{1\linewidth}{!}{
    \begin{tabular}{lccccccccc}
        \toprule
        \multirow{2}{*}{Method} & \multicolumn{4}{c}{Breakfast $\beta$ ($\alpha$ = 0.3)} & \multicolumn{4}{c}{50salads $\beta$ ($\alpha$ = 0.3)} \\
        \cmidrule(lr){2-5} \cmidrule(lr){6-9} 
        & 0.1 & 0.2 & 0.3 & 0.5 & 0.1 & 0.2 & 0.3 & 0.5\\
        \midrule
w$\backslash$o CRF & {35.09} & {31.41} & {29.34} & {27.20} & 31.13 & 27.09 & 22.43 & 17.92 \\
\rowcolor{myGray} Ours & \textbf{36.69} & \textbf{33.39} & \textbf{31.96} & \textbf{27.63} & \textbf{37.12} & \textbf{28.31} & \textbf{22.85} & \textbf{18.43} \\
         \bottomrule
    \end{tabular}}
    \caption{Impact of the CRF module on Breakfast and 50salads. }
    \label{tab:ablation_crf}
    \vspace{-6mm}
\end{table}
\begin{table}[t]
    \centering
    \setlength\tabcolsep{6pt}
    \resizebox{0.8\linewidth}{!}{
    \begin{tabular}{lcccc}
        \toprule
        \multirow{2}{*}{\shortstack{Init $\bm{M}$}} & \multicolumn{4}{c}{$\beta$ ($\alpha$ = 0.3)} \\
        \cmidrule(lr){2-5} 
        & 0.1 & 0.2 & 0.3 & 0.5 \\
        \midrule
        Pre-computed & \textbf{36.84} & {32.77} & {31.79} & {27.13}   \\
        \rowcolor{myGray} Random & {36.69} & \textbf{33.39} & \textbf{31.96} & \textbf{27.63}  \\
         \bottomrule
    \end{tabular}}
    \caption{Impact of matrix initialization for the Breakfast dataset.}
    \label{tab:ablation_matrix}
    \vspace{-6mm}
\end{table}

\paragraph{Global Temporal Sequence Optimization Loss.}
We also evaluate in 50salads and Breakfast the impact of using CRF layer at the top of decoder compared with removing it and training with a cross-entropy for each action independently as previous methods \cite{cycleConsistency, FUTR, ANTICIPATR}. \cref{tab:ablation_crf} shows that using CRF benefits the model in all the anticipation ratios demonstrating that decoding the most probable global sequence instead of individual action makes the model have a higher level of consistency.
Furthermore, as an alternative to a uniformly random matrix as initialization, we used also a pre-computed matrix based on the pairwise probability of actions, modeled as $P(a_{next}/a_{current})$, similar to the ETM no-decay matrix \cite{ETM}. The numerical results (see \cref{tab:ablation_matrix}) were quite similar, with pre-computed results slightly worse than the random initialization, despite pre-computed providing valuable knowledge. However, as demonstrated in \cref{fig:matrix}, the learned matrices exhibit significant similarities regardless of the initial conditions.

\subsection{Qualitative Results}
We further analyze qualitative results and compare them to those of FUTR \cite{FUTR}. In \cref{fig:qualitative}, we show the action segmentation at inference time in the observation interval and the predicted action segments in the anticipation interval.  
\cref{fig:qua_a} demonstrates that our model's accurate observation understanding, even with low $\alpha$ values in the Breakfast dataset, results in superior anticipation performance thanks to the encoder. Conversely, the action segmentation errors in the FUTR model leads to action anticipation mistakes. \cref{fig:qua_b} provides an example from the 50Salads dataset, where inter-class relations in anticipation are crucial. FUTR fails to model these relations accurately, producing prediction errors.
In contrast, our method, with improved temporal context consistency, effectively understands these relationships, ensuring plausible action transitions.

\section{Conclusions}
We introduced TCCA, a new transformer based encoder-decoder architecture for the task of LTA from untrimmed videos. Our method aims at ensuring quality predictions from observations as well as local and global temporal context consistency at inference time through the introduction of two dedicated modules, BACR aimed at maintaining context over long-term predictions and CRF, aimed at reducing logical errors at the output.
Exhaustive experiments and qualitative analysis over four benchmark datasets demonstrated the effectiveness of our approach over state-of-the-art methods and validated our algorithmic choices, demonstrating their individual benefit. Our approach outperforms probabilistic methods as well as methods based on LLM, and establishes state-of-the-art or remarkably competitive results.

\section{Acknowledgments}
This work was supported by projects PID2019-110977GA-I00 and PID2023-151351NB-I00 funded by MCIN/ AEI /10.13039/501100011033 and by ERDF, UE.
\bibliographystyle{plain} 

\bibliography{main}

\begin{thebibliography}{10}

\bibitem{abu2019uncertainty}
Yazan Abu~Farha and Juergen Gall.
\newblock Uncertainty-aware anticipation of activities.
\newblock In {\em Proceedings of the IEEE/CVF International Conference on Computer Vision Workshops}, pages 0--0, 2019.

\bibitem{cycleConsistency}
Yazan Abu~Farha, Qiuhong Ke, Bernt Schiele, and Juergen Gall.
\newblock Long-term anticipation of activities with cycle consistency.
\newblock In {\em Pattern Recognition: 42nd DAGM German Conference, DAGM GCPR 2020, T{\"u}bingen, Germany, September 28--October 1, 2020, Proceedings 42}, pages 159--173. Springer, 2021.

\bibitem{farhaWhenWillYou2018}
Yazan Abu~Farha, Alexander Richard, and Juergen Gall.
\newblock When will you do what?-anticipating temporal occurrences of activities.
\newblock In {\em Proceedings of the IEEE/CVF Conference on Computer Vision and Pattern Recognition}, pages 5343--5352, 2018.

\bibitem{LTContext}
Emad Bahrami, Gianpiero Francesca, and Juergen Gall.
\newblock How much temporal long-term context is needed for action segmentation?
\newblock In {\em Proceedings of the IEEE/CVF International Conference on Computer Vision}, pages 10351--10361, 2023.

\bibitem{UVAST}
Nadine Behrmann, S~Alireza Golestaneh, Zico Kolter, J{\"u}rgen Gall, and Mehdi Noroozi.
\newblock Unified fully and timestamp supervised temporal action segmentation via sequence to sequence translation.
\newblock In {\em Proceedings of the European Conference on Computer Vision}, pages 52--68. Springer, 2022.

\bibitem{brown2020language}
Tom Brown, Benjamin Mann, Nick Ryder, Melanie Subbiah, Jared~D Kaplan, Prafulla Dhariwal, Arvind Neelakantan, Pranav Shyam, Girish Sastry, Amanda Askell, et~al.
\newblock Language models are few-shot learners.
\newblock {\em Advances in Neural Information Processing Systems}, 33:1877--1901, 2020.

\bibitem{DETR}
Nicolas Carion, Francisco Massa, Gabriel Synnaeve, Nicolas Usunier, Alexander Kirillov, and Sergey Zagoruyko.
\newblock End-to-end object detection with transformers.
\newblock In {\em Proceedings of the European Conference on Computer Vision}, pages 213--229. Springer, 2020.

\bibitem{I3D}
Joao Carreira and Andrew Zisserman.
\newblock Quo vadis, action recognition? a new model and the kinetics dataset.
\newblock In {\em Proceedings of the IEEE/CVF Conference on Computer Vision and Pattern Recognition}, pages 6299--6308, 2017.

\bibitem{TASsurvey}
Guodong Ding, Fadime Sener, and Angela Yao.
\newblock Temporal action segmentation: An analysis of modern techniques.
\newblock {\em IEEE Transactions on Pattern Analysis and Machine Intelligence}, 2023.

\bibitem{MSTCN}
Yazan~Abu Farha and Jurgen Gall.
\newblock Ms-tcn: Multi-stage temporal convolutional network for action segmentation.
\newblock In {\em Proceedings of the IEEE/CVF Conference on Computer Vision and Pattern Recognition}, pages 3575--3584, 2019.

\bibitem{viterbi}
G.D. Forney.
\newblock The viterbi algorithm.
\newblock {\em Proceedings of the IEEE}, 61(3):268--278, 1973.

\bibitem{ETM}
Camilo~L Fosco, SouYoung Jin, Emilie Josephs, and Aude Oliva.
\newblock Leveraging temporal context in low representational power regimes.
\newblock In {\em Proceedings of the IEEE/CVF Conference on Computer Vision and Pattern Recognition}, pages 10693--10703, 2023.

\bibitem{gammulle2019forecasting}
Harshala Gammulle, Simon Denman, Sridha Sridharan, and Clinton Fookes.
\newblock Forecasting future action sequences with neural memory networks.
\newblock In {\em Proceedings of the British Machine Vision Conference}, 2019.

\bibitem{AVT}
Rohit Girdhar and Kristen Grauman.
\newblock Anticipative video transformer.
\newblock In {\em Proceedings of the IEEE/CVF International Conference on Computer Vision}, pages 13505--13515, October 2021.

\bibitem{girdhar2017actionvlad}
Rohit Girdhar, Deva Ramanan, Abhinav Gupta, Josef Sivic, and Bryan Russell.
\newblock Actionvlad: Learning spatio-temporal aggregation for action classification.
\newblock In {\em Proceedings of the IEEE/CVF Conference on Computer Vision and Pattern Recognition}, pages 971--980, 2017.

\bibitem{gong2024actfusion}
Dayoung Gong, Suha Kwak, and Minsu Cho.
\newblock Actfusion: a unified diffusion model for action segmentation and anticipation.
\newblock {\em Advances in Neural Information Processing Systems}, 2024.

\bibitem{gong2024activity}
Dayoung Gong, Joonseok Lee, Deunsol Jung, Suha Kwak, and Minsu Cho.
\newblock Activity grammars for temporal action segmentation.
\newblock {\em Advances in Neural Information Processing Systems}, 36, 2024.

\bibitem{FUTR}
Dayoung Gong, Joonseok Lee, Manjin Kim, Seong~Jong Ha, and Minsu Cho.
\newblock Future transformer for long-term action anticipation.
\newblock In {\em Proceedings of the IEEE/CVF Conference on Computer Vision and Pattern Recognition}, pages 3052--3061, 2022.

\bibitem{gupta2022act}
Akash Gupta, Jingen Liu, Liefeng Bo, Amit~K Roy-Chowdhury, and Tao Mei.
\newblock A-act: Action anticipation through cycle transformations.
\newblock {\em arXiv preprint arXiv:2204.00942}, 2022.

\bibitem{GELU}
Dan Hendrycks and Kevin Gimpel.
\newblock Gaussian error linear units (gelus), 2023.

\bibitem{huang2015bidirectionallstmcrfmodelssequence}
Zhiheng Huang, Wei Xu, and Kai Yu.
\newblock Bidirectional lstm-crf models for sequence tagging, 2015.

\bibitem{hussein2019timeception}
Noureldien Hussein, Efstratios Gavves, and Arnold~WM Smeulders.
\newblock Timeception for complex action recognition.
\newblock In {\em Proceedings of the IEEE/CVF Conference on Computer Vision and Pattern Recognition}, pages 254--263, 2019.

\bibitem{hussein2019videograph}
Noureldien Hussein, Efstratios Gavves, and Arnold~WM Smeulders.
\newblock Videograph: Recognizing minutes-long human activities in videos.
\newblock In {\em ICCV Workshop}, 2019.

\bibitem{keTimeConditionedActionAnticipation2019}
Qiuhong Ke, Mario Fritz, and Bernt Schiele.
\newblock Time-{{Conditioned Action Anticipation}} in {{One Shot}}.
\newblock In {\em Proceedings of the IEEE/CVF Conference on Computer Vision and Pattern Recognition}, June 2019.

\bibitem{LALM}
Sanghwan Kim, Daoji Huang, Yongqin Xian, Otmar Hilliges, Luc~Van Gool, and Xi~Wang.
\newblock Palm: Predicting actions through language models, 2024.

\bibitem{breakfast}
Hilde Kuehne, Ali Arslan, and Thomas Serre.
\newblock The language of actions: Recovering the syntax and semantics of goal-directed human activities.
\newblock In {\em Proceedings of the IEEE/CVF Conference on Computer Vision and Pattern Recognition}, pages 780--787, 2014.

\bibitem{kuehne2016end}
Hilde Kuehne, Juergen Gall, and Thomas Serre.
\newblock An end-to-end generative framework for video segmentation and recognition.
\newblock In {\em Proceedings of the IEEE/CVF Winter Conference on Applications of Computer Vision}, pages 1--8, 2016.

\bibitem{kuehne2018hybrid}
Hilde Kuehne, Alexander Richard, and Juergen Gall.
\newblock A hybrid rnn-hmm approach for weakly supervised temporal action segmentation.
\newblock {\em IEEE Transactions on Pattern Analysis and Machine Intelligence}, 42(4):765--779, 2018.

\bibitem{kullback1951information}
Solomon Kullback and Richard~A Leibler.
\newblock On information and sufficiency.
\newblock {\em The annals of mathematical statistics}, 22(1):79--86, 1951.

\bibitem{CRF}
John Lafferty, Andrew McCallum, Fernando Pereira, et~al.
\newblock Conditional random fields: Probabilistic models for segmenting and labeling sequence data.
\newblock In {\em Icml}, volume~1, page~3. Williamstown, MA, 2001.

\bibitem{lea2017temporal}
Colin Lea, Michael~D Flynn, Rene Vidal, Austin Reiter, and Gregory~D Hager.
\newblock Temporal convolutional networks for action segmentation and detection.
\newblock In {\em Proceedings of the IEEE/CVF Conference on Computer Vision and Pattern Recognition}, pages 156--165, 2017.

\bibitem{li2020ms}
Shijie Li, Yazan~Abu Farha, Yun Liu, Ming-Ming Cheng, and Juergen Gall.
\newblock Ms-tcn++: Multi-stage temporal convolutional network for action segmentation.
\newblock {\em IEEE Transactions on Pattern Analysis and Machine Intelligence}, 45(6):6647--6658, 2020.

\bibitem{EGTEA}
Yin Li, Miao Liu, and James~M Rehg.
\newblock In the eye of beholder: Joint learning of gaze and actions in first person video.
\newblock In {\em Proceedings of the European Conference on Computer Vision}, pages 619--635, 2018.

\bibitem{WARMPUP}
Ilya Loshchilov and Frank Hutter.
\newblock {SGDR:} stochastic gradient descent with warm restarts.
\newblock In {\em Proceedings of the International Conference on Learning Representations}, 2017.

\bibitem{ADAMW}
Ilya Loshchilov and Frank Hutter.
\newblock Decoupled weight decay regularization.
\newblock In {\em International Conference on Learning Representations}, 2019.

\bibitem{FACT}
Zijia Lu and Ehsan Elhamifar.
\newblock Fact: Frame-action cross-attention temporal modeling for efficient action segmentation.
\newblock In {\em Proceedings of the IEEE/CVF Conference on Computer Vision and Pattern Recognition}, pages 18175--18185, June 2024.

\bibitem{mavroudi2018end}
Effrosyni Mavroudi, Divya Bhaskara, Shahin Sefati, Haider Ali, and Ren{\'e} Vidal.
\newblock End-to-end fine-grained action segmentation and recognition using conditional random field models and discriminative sparse coding.
\newblock In {\em Proceedings of the IEEE/CVF Winter Conference on Applications of Computer Vision}, pages 1558--1567, 2018.

\bibitem{mittal2024can}
Himangi Mittal, Nakul Agarwal, Shao-Yuan Lo, and Kwonjoon Lee.
\newblock Can't make an omelette without breaking some eggs: Plausible action anticipation using large video-language models.
\newblock In {\em Proceedings of the IEEE/CVF Conference on Computer Vision and Pattern Recognition}, pages 18580--18590, 2024.

\bibitem{EGOTOPO}
Tushar Nagarajan, Yanghao Li, Christoph Feichtenhofer, and Kristen Grauman.
\newblock Ego-topo: Environment affordances from egocentric video.
\newblock In {\em Proceedings of the IEEE/CVF Conference on Computer Vision and Pattern Recognition}, pages 163--172, 2020.

\bibitem{ANTICIPATR}
Megha Nawhal, Akash~Abdu Jyothi, and Greg Mori.
\newblock Rethinking learning approaches for long-term action anticipation.
\newblock In {\em Proceedings of the European Conference on Computer Vision}, pages 558--576. Springer, 2022.

\bibitem{senerTemporalAggregateRepresentations2020}
Fadime Sener, Dipika Singhania, and Angela Yao.
\newblock Temporal aggregate representations for long-range video understanding.
\newblock In {\em Proceedings of the European Conference on Computer Vision}, pages 154--171. Springer, 2020.

\bibitem{50salads}
Sebastian Stein and Stephen~J McKenna.
\newblock Combining embedded accelerometers with computer vision for recognizing food preparation activities.
\newblock In {\em Proceedings of the 2013 ACM International Joint Conference on Pervasive and Ubiquitous Computing}, pages 729--738, 2013.

\bibitem{vaswani2017attention}
Ashish Vaswani, Noam Shazeer, Niki Parmar, Jakob Uszkoreit, Llion Jones, Aidan~N Gomez, {\L}ukasz Kaiser, and Illia Polosukhin.
\newblock Attention is all you need.
\newblock {\em Advances in Neural Information Processing Systems}, 30, 2017.

\bibitem{wang2023vamos}
Shijie Wang, Qi~Zhao, Minh~Quan Do, Nakul Agarwal, Kwonjoon Lee, and Chen Sun.
\newblock Vamos: Versatile action models for video understanding.
\newblock In {\em Proceedings of the European Conference on Computer Vision}, 2023.

\bibitem{xu2022don}
Ziwei Xu, Yogesh Rawat, Yongkang Wong, Mohan~S Kankanhalli, and Mubarak Shah.
\newblock Don't pour cereal into coffee: Differentiable temporal logic for temporal action segmentation.
\newblock {\em Advances in Neural Information Processing Systems}, 35:14890--14903, 2022.

\bibitem{gtda2024zatsarynna}
Olga Zatsarynna, Emad Bahrami, Yazan~Abu Farha, Gianpiero Francesca, and Juergen Gall.
\newblock Gated temporal diffusion for stochastic long-term dense anticipation.
\newblock In {\em Proceedings of the European Conference on Computer Vision}, 2024.

\bibitem{zhang2024object}
Ce~Zhang, Changcheng Fu, Shijie Wang, Nakul Agarwal, Kwonjoon Lee, Chiho Choi, and Chen Sun.
\newblock Object-centric video representation for long-term action anticipation.
\newblock In {\em Proceedings of the IEEE/CVF Winter Conference on Applications of Computer Vision}, pages 6751--6761, 2024.

\bibitem{zhao2020diverse}
He~Zhao and Richard~P Wildes.
\newblock On diverse asynchronous activity anticipation.
\newblock In {\em Proceedings of the European Conference on Computer Vision}, pages 781--799. Springer, 2020.

\bibitem{ANTGPT}
Qi~Zhao, Shijie Wang, Ce~Zhang, Changcheng Fu, Minh~Quan Do, Nakul Agarwal, Kwonjoon Lee, and Chen Sun.
\newblock Antgpt: Can large language models help long-term action anticipation from videos?, 2024.

\bibitem{ANTSurvey}
Zeyun Zhong, Manuel Martin, Michael Voit, Juergen Gall, and Jürgen Beyerer.
\newblock A survey on deep learning techniques for action anticipation, 2023.

\bibitem{DIFFANT}
Zeyun Zhong, Chengzhi Wu, Manuel Martin, Michael Voit, Juergen Gall, and J{\"u}rgen Beyerer.
\newblock Diffant: Diffusion models for action anticipation.
\newblock {\em arXiv preprint arXiv:2311.15991}, 2023.

\end{thebibliography}

\clearpage
\setcounter{page}{1}

\section{Additional Ablation Study}
We have extended the ablation study presented in \Cref{sec:ablation} to include all datasets: Breakfast, 50Salads, EGTEA, and EpicKitchens-55 (see \Cref{tab:ablation_sup}). \Cref{tab:sup_ablation_segmentation_loss} highlights the importance of using the smoothing loss to enhance anticipation metrics by reducing segmentation noise. \Cref{tab:sup_ablation_encoder} demonstrates that although FACT may achieve better segmentation of past observations for some datasets, LTContext consistently yields superior results. Furthermore, \Cref{tab:supp_ablation_BACR} and \Cref{tab:supp_ablation_crf} illustrate that our proposed decoder modules lead to higher performance across all datasets.

\section{Additional Details}
\paragraph{Duration architecture.}

As explained in \cref{sec:method_decoder}, the duration is computed based on both, the action query $\bm{Q^\prime}$ and the action logits $\hat{{a}}_{pres}$. In \cref{tab:ablation_duration}, we also report the performance  using a separated head for the duration that only rely on $\bm{Q^\prime}$. The results indicate a slight performance increase with the first approach. This improvement is likely because the decoder can better model the intrinsic action duration of several action in the dataset. 

\begin{figure*}
  \vspace{-5mm}

  \centering
  \begin{subfigure}[t]{0.6\linewidth}
        \vspace*{0pt} 

      \includegraphics[width=1\linewidth]{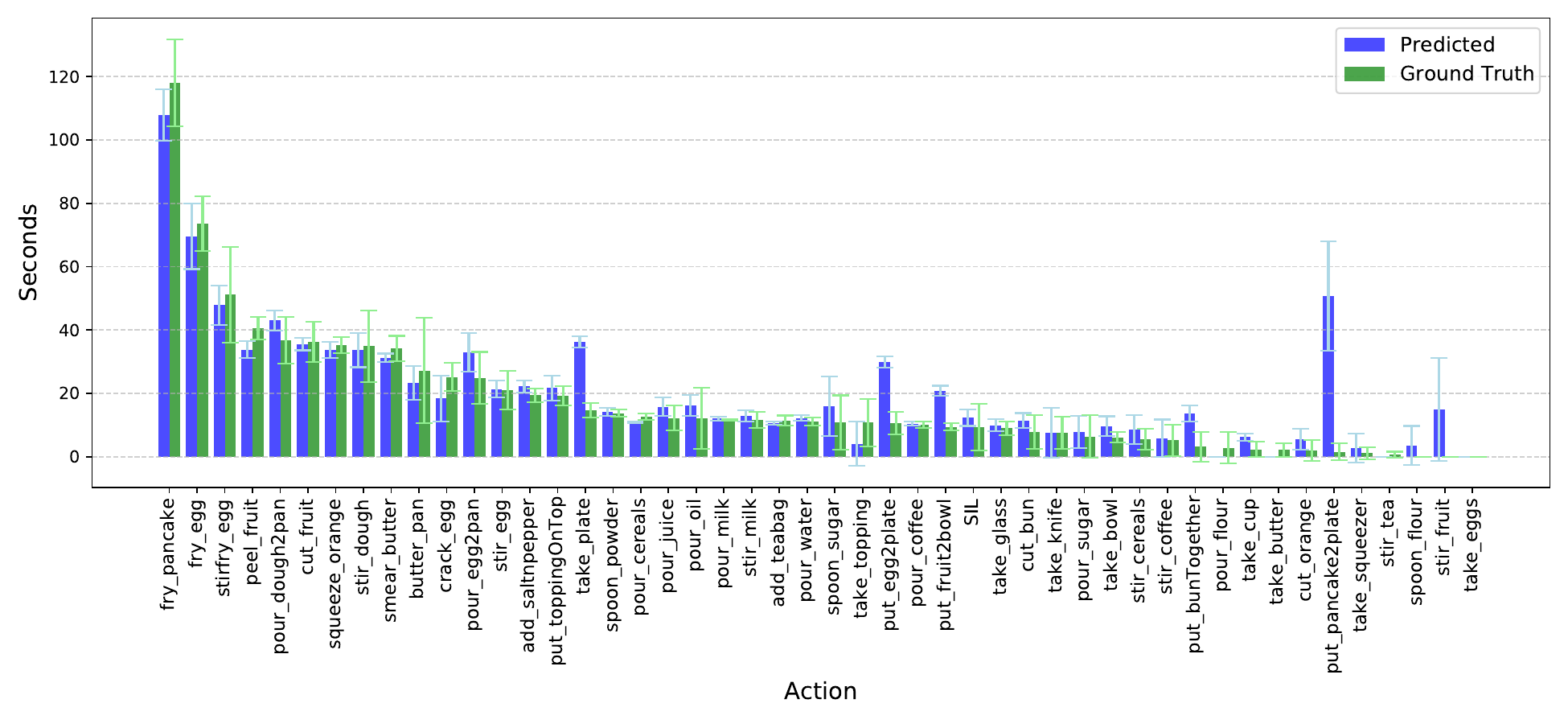}
    \caption{Breakfast}
    \label{fig:duration-breakfast}
    \vspace{-2mm}

  \end{subfigure}
  \hfill
  \begin{subfigure}[t]{0.38\linewidth}
      \vspace*{0pt} 

    \includegraphics[width=1\linewidth]{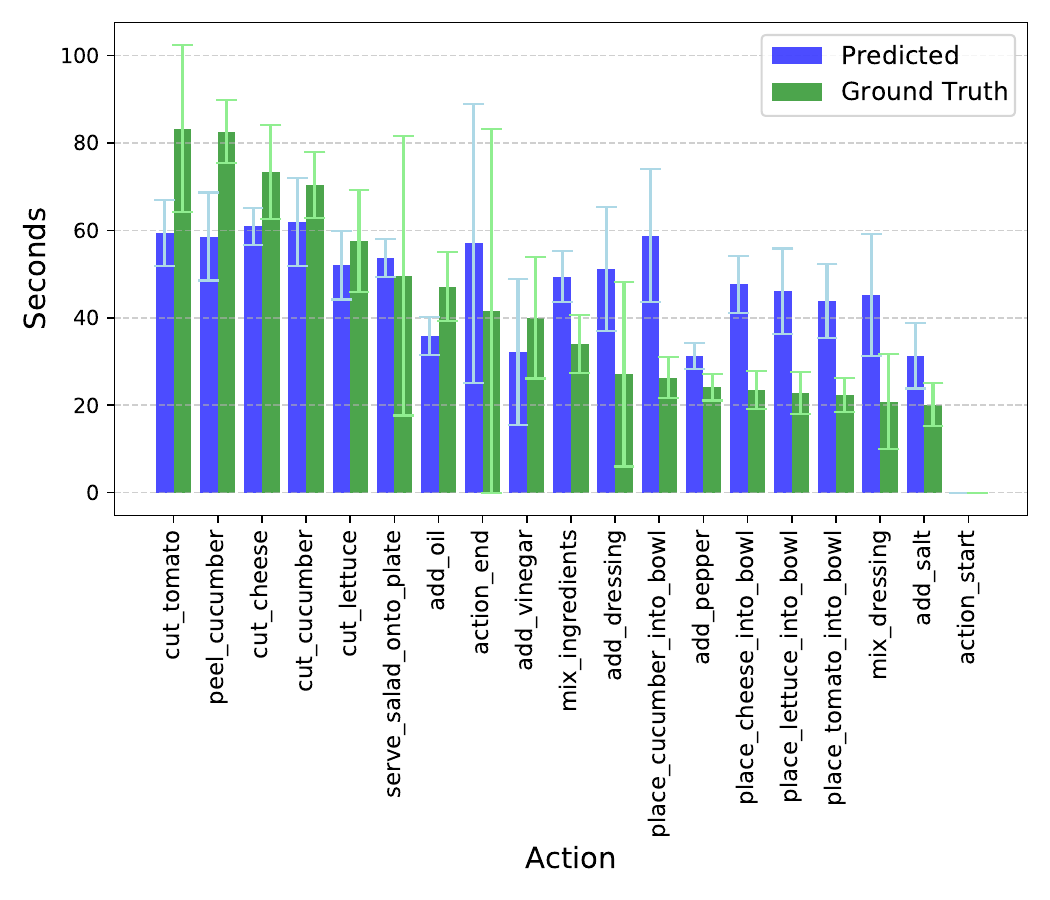}    
    \caption{50salads}
    \label{fig:duration-50salads}
  \vspace{-2mm}
  \end{subfigure}

  \caption{\textbf{Duration analysis for Breakfast and 50salads. } Average duration per action for both ground-truth (green) and predicted (blue). Actions ordered by the ground-truth duration.}
  \label{fig:duration_plot}
  \vspace{-3mm}
\end{figure*}

We conducted a detailed analysis of the duration prediction for the Breakfast and 50Salads datasets to evaluate its performance. We calculated the ground-truth mean and standard deviation in seconds for the future segments ($\alpha=0.3$) grouped by each class, and compared these against the predicted durations provided by our model for each class. For the Breakfast dataset (\cref{fig:duration-breakfast}), the model effectively captured the duration for more frequent classes. However, the error was higher for certain infrequent classes, likely due to insufficient examples in the training data. Conversely, the duration modeling for the 50Salads dataset (\cref{fig:duration-50salads}) proved more challenging. Regardless of whether the actions were long or short, the model tended to predict similar durations, ranging from 40 to 60 seconds. We hypothesize that this limitation arises from the small dataset size and the lack of sufficient training examples.

\paragraph{Number of queries.}
We also analyze the number of queries in the Breakfast dataset, comparing query counts of 6, 8, and 10 (see \cref{tab:ablation_queries}). Our findings indicate that the optimal number of queries for our method is 8, as variations in the number of queries—either more or less—do not significantly improve the results. 

\begin{figure}
  \centering
  \begin{subfigure}{0.49\linewidth}
      \includegraphics[width=1\linewidth]{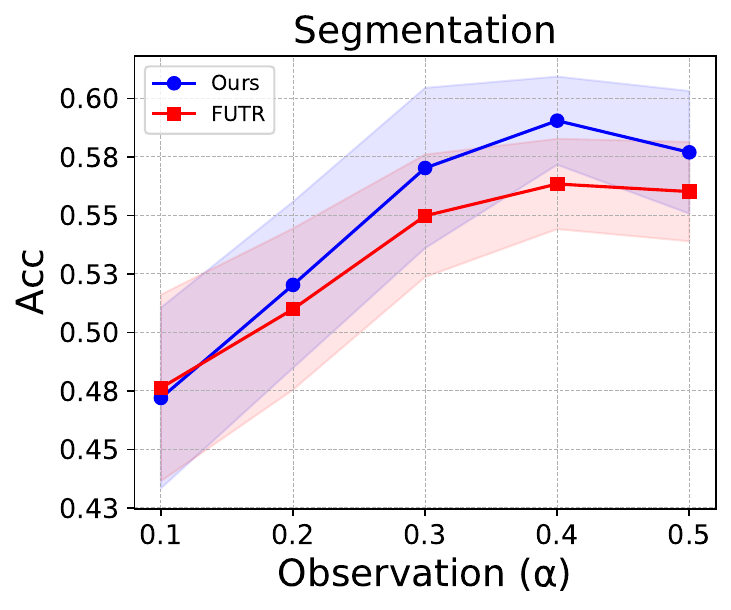}
    \caption{Segmentation}
    \label{fig:analysis_Segmentation}
  \end{subfigure}
  \hfill
  \begin{subfigure}{0.49\linewidth}
    \includegraphics[width=1\linewidth]{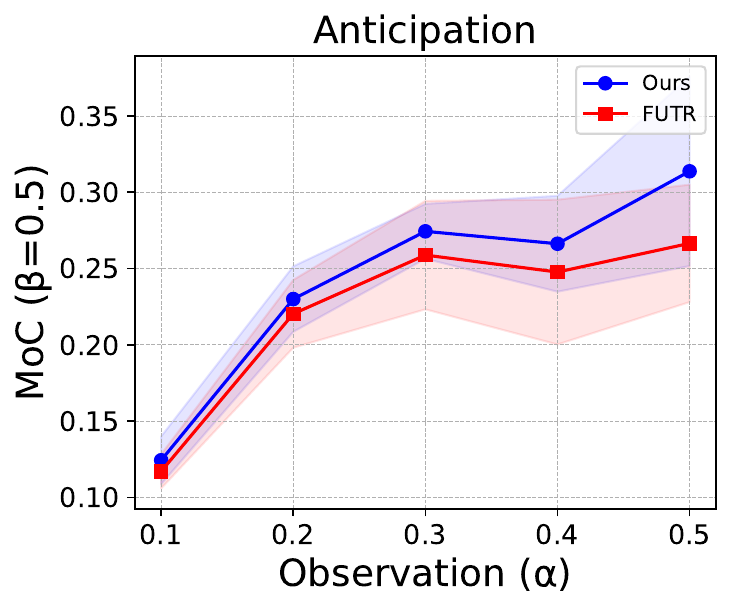}    
    \caption{Anticipation}
    \label{fig:analysis_Anticipation}
  \end{subfigure}
  \caption{Impact of Observation Horizon ($\alpha$) and Segmentation on anticipation in Breakfast. Results are shown for $\alpha \in \{0.1, 0.2, 0.3, 0.4, 0.5\}$. Segmentation is measured with Acc and Anticipation with MoC at $\beta=0.5$. Points indicate the mean; shaded areas represent the standard deviation across 4 splits.}
  \label{fig:analysis}
  \vspace{-4mm}
\end{figure}
\begin{table*}[t]
    \centering
    \setlength\tabcolsep{2pt}

    \begin{subtable}[t]{\linewidth}
        \centering
    \resizebox{1\linewidth}{!}{\begin{tabular}{@{}cccccccccccccccccc@{}}
        \toprule
        \multicolumn{2}{c}{Loss}  & \multicolumn{4}{c}{Breakfast} & \multicolumn{4}{c}{50salads} & \multicolumn{4}{c}{EGTEA Gaze+} & \multicolumn{4}{c}{EpicKitchens-55} \\

        \cmidrule(lr){1-2} \cmidrule(lr){3-6} \cmidrule(lr){7-10} \cmidrule(lr){11-14} \cmidrule(lr){15-18} 
        $\mathcal{L}_\text{seg}$ & $\mathcal{L}_\text{s}$ & mMoC  & Acc & Edit & F1@50 & mMoC  & Acc & Edit & F1@50 & All  & Acc & Edit & F1@50 & All  & Acc & Edit & F1@50\\
        \midrule
          &  & {25.51} & - & - & - & 16.84 & - & - & - & 75.3 & - & - & - & 32.2 & - & - & - \\
\checkmark &  & \underline{27.95} & \underline{54.37} & \underline{60.57} & \underline{41.70} & \underline{25.67} & \underline{69.13} & \underline{61.71} & \underline{51.77} & \underline{77.4} & \textbf{49.9} & \textbf{47.5} & \textbf{25.1} & \underline{38.4} & \textbf{43.1} & \underline{39.9} & \textbf{26.5} \\
\rowcolor{myGray} \checkmark & \checkmark & \textbf{29.13} & \textbf{54.53} & \textbf{62.01} & \textbf{41.86} & \textbf{28.78} & \textbf{70.75} & \textbf{62.16} & \textbf{54.51} & \textbf{80.3} & \underline{46.4} & \underline{45.8} & \underline{22.1} & \textbf{40.6} & \underline{42.2} & \textbf{40.7} & \underline{26.4} \\
         \bottomrule
    \end{tabular}}
    \caption{Segmentation loss.}
    \label{tab:sup_ablation_segmentation_loss}
    \vspace{2mm}
    \end{subtable}
    \begin{subtable}[t]{\linewidth}
        \centering
    \resizebox{1\linewidth}{!}{\begin{tabular}{@{}lcccccccccccccccc@{}}
        \toprule
        \multirow{2}{*}{Encoder}  & \multicolumn{4}{c}{Breakfast} & \multicolumn{4}{c}{50salads} & \multicolumn{4}{c}{EGTEA Gaze+} & \multicolumn{4}{c}{EpicKitchens-55} \\

         \cmidrule(lr){2-5} \cmidrule(lr){6-9} \cmidrule(lr){10-13} \cmidrule(lr){14-17} 
         & mMoC  & Acc & Edit & F1@50 & mMoC  & Acc & Edit & F1@50 & All  & Acc & Edit & F1@50 & All  & Acc & Edit & F1@50\\
        \midrule
DETR \cite{DETR} & \underline{25.94} & 53.30 & 29.30 & 23.28 & \underline{26.95} & \underline{67.05} & 3.99 & 2.99 & \underline{76.7} & \underline{48.0} & 30.0 & 18.7 & \underline{35.1} & 40.4 & 27.4 & 17.6 \\
FACT \cite{FACT} & 23.40 & \underline{53.70} & \underline{51.37} & \underline{37.86} & 20.17 & 66.31 & \underline{28.71} & \underline{24.72} & 69.4 & \textbf{51.3} & \underline{45.6} & \textbf{27.5} & 26.9 & \textbf{42.6} & \underline{39.8} & \underline{22.5} \\
\rowcolor{myGray} LTContext \cite{LTContext} & \textbf{29.13} & \textbf{54.53} & \textbf{62.01} & \textbf{41.86} & \textbf{28.78} & \textbf{70.75} & \textbf{62.16} & \textbf{54.51} & \textbf{80.3} & 46.4 & \textbf{45.8} & \underline{22.1} & \textbf{40.6} & \underline{42.2} & \textbf{40.7} & \textbf{26.4} \\
         \bottomrule
    \end{tabular}}
    \caption{Encoder architecture.}
    \label{tab:sup_ablation_encoder}
    \vspace{2mm}
    \end{subtable}

    \begin{subtable}[t]{\linewidth}
    \setlength\tabcolsep{4pt}
    \centering
    \resizebox{1\linewidth}{!}{\begin{tabular}{@{}cccccccccccccccc@{}}
        \toprule
        \multicolumn{2}{c}{Loss} & \multicolumn{4}{c}{Breakfast $\beta$ ($\alpha$ = 0.3)} & \multicolumn{4}{c}{50salads $\beta$ ($\alpha$ = 0.3)} & \multicolumn{3}{c}{EGTEA Gaze+} & \multicolumn{3}{c}{EpicKitchens-55}\\
        \cmidrule(lr){1-2} \cmidrule(lr){3-6} \cmidrule(lr){7-10} \cmidrule(lr){11-13} \cmidrule(lr){14-16} 
        $\mathcal{L}_\text{prev}$ & $\mathcal{L}_\text{next}$  & 0.1 & 0.2 & 0.3 & 0.5 & 0.1 & 0.2 & 0.3 & 0.5 & All & Freq & Rare & All & Freq & Rare  \\
        \midrule
 &  & \underline{35.52} & 31.65 & 29.99 & 26.63 & 27.16 & 23.97 & 19.26 & 15.70 & 74.3 & 80.7 & 57.8 & 37.5 & 57.2 & 26.9 \\
 & \checkmark & 35.11 & 30.99 & 29.87 & 26.33 & 29.94 & 23.84 & 21.79 & 16.05 & \underline{76.5} & \underline{82.4} & 61.1 & \underline{40.0} & \underline{58.6} & \underline{30.2} \\
\checkmark &  & 35.20 & \underline{31.96} & \underline{30.60} & \underline{26.86} & \underline{30.40} & \underline{26.69} & \underline{22.55} & \textbf{18.65} & 74.1 & 77.8 & \underline{64.4} & 35.5 & 58.2 & 23.3 \\
\rowcolor{myGray} \checkmark & \checkmark & \textbf{36.69} & \textbf{33.39} & \textbf{31.96} & \textbf{27.63} & \textbf{37.12} & \textbf{28.31} & \textbf{22.85} & \underline{18.43} & \textbf{80.3} & \textbf{85.8} & \textbf{65.7} & \textbf{40.6} & \textbf{59.4} & \textbf{30.7} \\
         \bottomrule
    \end{tabular}}
    \caption{BACR module.}
    \label{tab:supp_ablation_BACR}
    \vspace{2mm}
\end{subtable}

    \begin{subtable}[t]{\linewidth}
        \setlength\tabcolsep{4pt}

        \centering
        \resizebox{1\linewidth}{!}{\begin{tabular}{lcccccccccccccc}
            \toprule
            \multirow{2}{*}{Method} & \multicolumn{4}{c}{Breakfast $\beta$ ($\alpha$ = 0.3)} & \multicolumn{4}{c}{50salads $\beta$ ($\alpha$ = 0.3)}  & \multicolumn{3}{c}{EGTEA Gaze+} & \multicolumn{3}{c}{EpicKitchens-55}\\
            \cmidrule(lr){2-5} \cmidrule(lr){6-9}  \cmidrule(lr){10-12} \cmidrule(lr){13-15} 
            & 0.1 & 0.2 & 0.3 & 0.5 & 0.1 & 0.2 & 0.3 & 0.5 &  All & Freq & Rare &  All & Freq & Rare\\
            \midrule
    w$\backslash$o CRF & {35.09} & {31.41} & {29.34} & {27.20} & 31.13 & 27.09 & 22.43 & 17.92 &  79.9 & 85.1 & 64.7 & 39.6 & 57.6 & 30.0 \\
    \rowcolor{myGray} Ours & \textbf{36.69} & \textbf{33.39} & \textbf{31.96} & \textbf{27.63}  & \textbf{37.12} & \textbf{28.31} & \textbf{22.85} & \textbf{18.43} & \textbf{80.3} & \textbf{85.8} & \textbf{65.7} & \textbf{40.6} & \textbf{59.4} & \textbf{30.7} \\
             \bottomrule
        \end{tabular}}
        \caption{Global temporal sequence optimization}
        \label{tab:supp_ablation_crf}
        \vspace{-2mm}
    \end{subtable}

    \caption{\textbf{Additional Ablation in All Datasets} \cref{tab:sup_ablation_segmentation_loss} and \cref{tab:sup_ablation_encoder} present the ablation results for segmentation loss and encoder architecture, respectively. For the datasets Breakfast and 50Salads, we report the anticipation metric mMOC (mean of MoC over all observed and predicted horizons). For EGTEA and EpicKitchens, we use the \textit{All} metric. Additionally, we include segmentation metrics for the past observation encoder (\textit{Acc, Edit, F1@50}). \cref{tab:supp_ablation_BACR} and \cref{tab:supp_ablation_crf} display the ablation results for the BACR and Global Temporal Sequence Optimization, respectively. For Breakfast and 50Salads, we report all prediction horizons with $\alpha=0.3$. For EGTEA and EpicKitchens, we report all metrics (\textit{All, Freq, Rare}).} 
    \label{tab:ablation_sup}
    \vspace{-1mm}

\end{table*}
\begin{table*}[t]
    \centering
    \begin{subtable}[t]{\linewidth}
        \centering
        \begin{tabular}{lccccccccc}
        \toprule
        \multirow{2}{*}{Duration} & \multicolumn{4}{c}{Breakfast $\beta$ ($\alpha$ = 0.2)} & \multicolumn{4}{c}{Breakfast $\beta$ ($\alpha$ = 0.3)} \\
        \cmidrule(lr){2-5} \cmidrule(lr){6-9}
        & 0.1 & 0.2 & 0.3 & 0.5 & 0.1 & 0.2 & 0.3 & 0.5 \\
        \midrule
Independent & 27.39 & {25.77} & 24.01 & 22.91 & \textbf{36.96} & {33.19} & 30.20 & 27.04 \\
\rowcolor{myGray} Partially dependent & \textbf{28.83} & \textbf{25.93} & \textbf{25.19} & \textbf{23.41} & {36.69} & \textbf{33.39} & \textbf{31.96} & \textbf{27.63} \\
         \bottomrule
    \end{tabular}
    \caption{Duration architecture.}
    \label{tab:ablation_duration}
    \vspace{1mm}

    \end{subtable}
    
    \begin{subtable}[t]{\linewidth}
        \centering
        \begin{tabular}{cccccccccc}
        \toprule
        \multirow{2}{*}{Queries} & \multicolumn{4}{c}{Breakfast $\beta$ ($\alpha$ = 0.2)} & \multicolumn{4}{c}{Breakfast $\beta$ ($\alpha$ = 0.3)} \\
        \cmidrule(lr){2-5} \cmidrule(lr){6-9} 
        & 0.1 & 0.2 & 0.3 & 0.5 & 0.1 & 0.2 & 0.3 & 0.5 \\
        \midrule
        6 & 27.01 & 24.78 & 24.04 & 22.68 & 34.13 & 30.65 & 29.05 & 25.65 \\
        \rowcolor{myGray} 8 & \textbf{28.83} & \textbf{25.93} & \textbf{25.19} & \textbf{23.41} & \textbf{36.69} & \textbf{33.39} & \textbf{31.96} & \underline{27.63} \\
        10 & \underline{27.85} & \underline{25.06} & \underline{24.10} & \underline{23.21} & \underline{35.86} & \underline{32.32} & \underline{31.16} & \textbf{27.74} \\

         \bottomrule
    \end{tabular}
    \caption{Number of action queries.}
    \label{tab:ablation_queries}
        \vspace{-2mm}
    \end{subtable}

    \caption{Additional analysis for Breakfast dataset}
    \label{tab:analysis_sup}
    \vspace{2mm}
\end{table*}
\paragraph{Effects of action segmentation on anticipation}
The goal of this experiment is to determine whether enhancing action segmentation improves anticipation and to understand how the observation horizon ($\alpha$) affects anticipation. As illustrated in \cref{fig:analysis_Anticipation}, the context is highly important for achieving good performance in action anticipation. Similarly, \cref{fig:analysis_Segmentation} shows that, as with anticipation, a larger context leads to better action segmentation. Moreover, our model outperforms FUTR in both segmentation and anticipation. Therefore, we can assure that improving action segmentation enhances anticipation for the Breakfast dataset.

\section{Additional Implementation Details}

\paragraph{Training}
For training, we use a different batch size and dropout rate for each dataset. Specifically, the batch sizes and dropout rates are as follows: 16 and 0.3 for Breakfast, 8 and 0.1 for 50salads, 4 and 0.2 for EpicKitchens-55, and 8 and 0.3 for EGTEA. Additionally, we apply a gradient clipping of 10 to stabilize training. All experiments were conducted on a single NVIDIA RTX 3090 GPU.
\paragraph{Complexity of Viterbi on inference}
 The complexity of Viterbi algorithm for the LTA task is $\mathcal{O}(TS^2)$, where $S$ and $T$ are the number of classes in the dataset and the number of actions to be predicted in the video (that is equal to the number of estimated segments), respectively. In 50Salads $S=17, T_ {avg} = 20$ and in Breakfast  $S=48, T_{avg}= 8$. Therefore the running time is much more reduced with respect to when it is applied for action segmentation, where $T$ corresponds to the number of frames.

\section{Additional Qualitative Results}
\begin{figure*}[t]
    \centering
    \begin{subfigure}[t]{\linewidth}
    \includegraphics[width=\columnwidth]{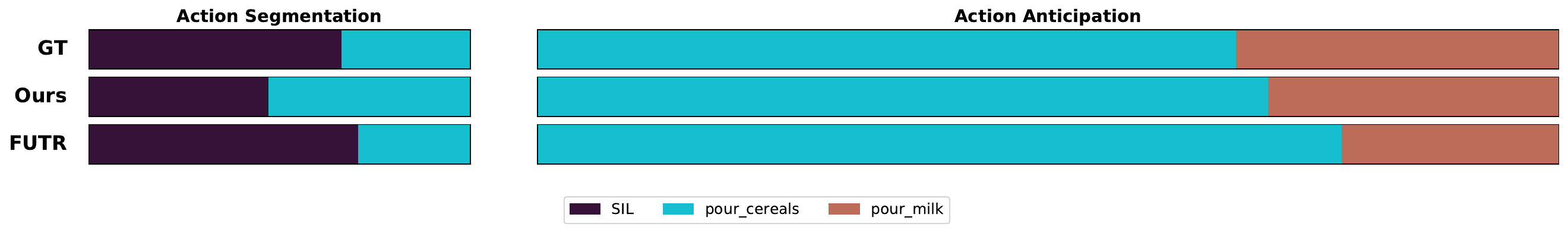}
    \caption{\textbf{Breakfast.} P29\_cam01\_P29\_cereals; ($\alpha$=0.2, $\beta$=0.5)}
    \label{fig:add_qua_a}
    \end{subfigure}
    \vspace{-1mm}
    \begin{subfigure}[t]{\linewidth}
    \includegraphics[width=\columnwidth]{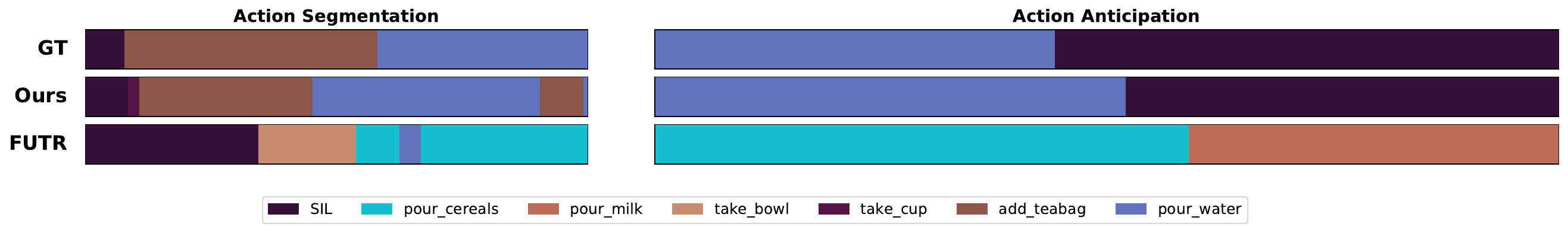}
    \caption{\textbf{Breakfast.} P36\_cam01\_P36\_tea; ($\alpha$=0.3, $\beta$=0.5)}
    \label{fig:add_qua_b}
    \end{subfigure}
    \vspace{-1mm}
    \begin{subfigure}[t]{\linewidth}
    \includegraphics[width=\columnwidth]{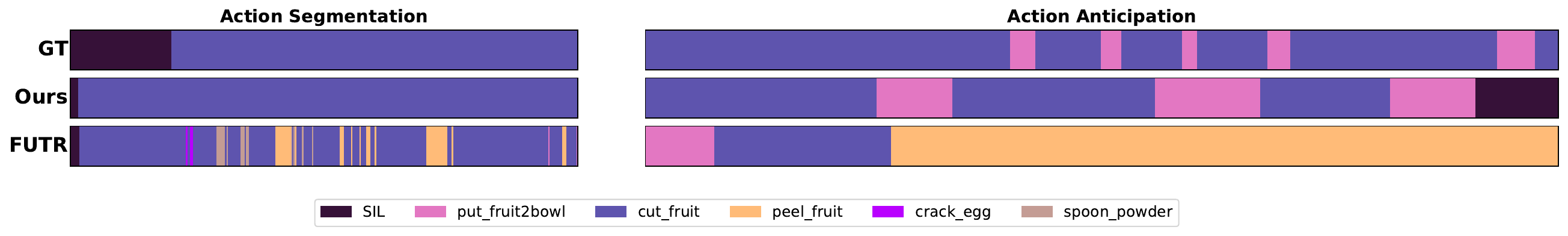}
    \caption{\textbf{Breakfast.} P32\_cam01\_P32\_salat; ($\alpha$=0.3, $\beta$=0.5)}
    \label{fig:add_qua_c}
    \end{subfigure}
    \vspace{-1mm}
     \begin{subfigure}[t]{\linewidth}
    \includegraphics[width=\columnwidth]{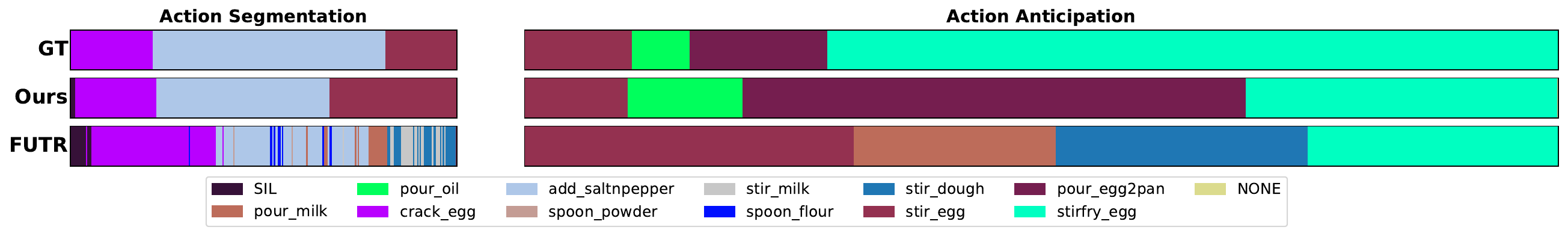}
    \caption{\textbf{Breakfast.} P34\_cam01\_P34\_scrambledegg; ($\alpha$=0.3, $\beta$=0.5)}
    \label{fig:add_qua_d}
    \end{subfigure}
    \vspace{-1mm}
     \begin{subfigure}[t]{\linewidth}
    \includegraphics[width=\columnwidth]{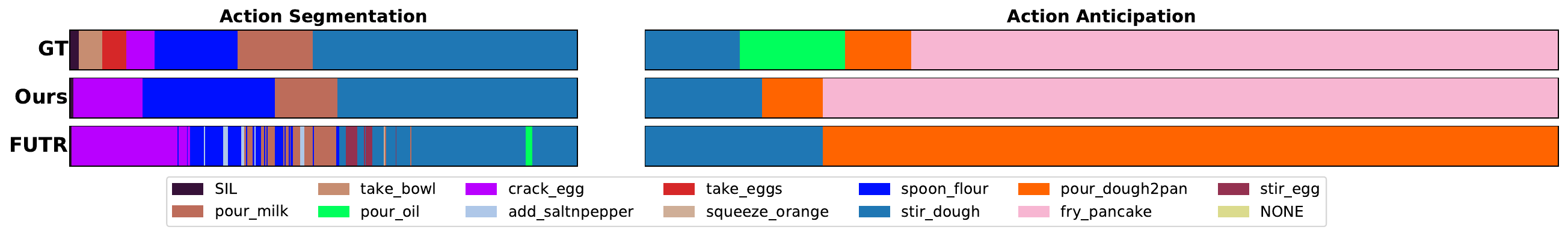}
    \caption{\textbf{Breakfast.} P22\_cam01\_P22\_pancake; ($\alpha$=0.3, $\beta$=0.5)}
    \label{fig:add_qua_e}
    \end{subfigure}
    \vspace{-1mm}
    \begin{subfigure}[t]{\linewidth}
    \includegraphics[width=\columnwidth]{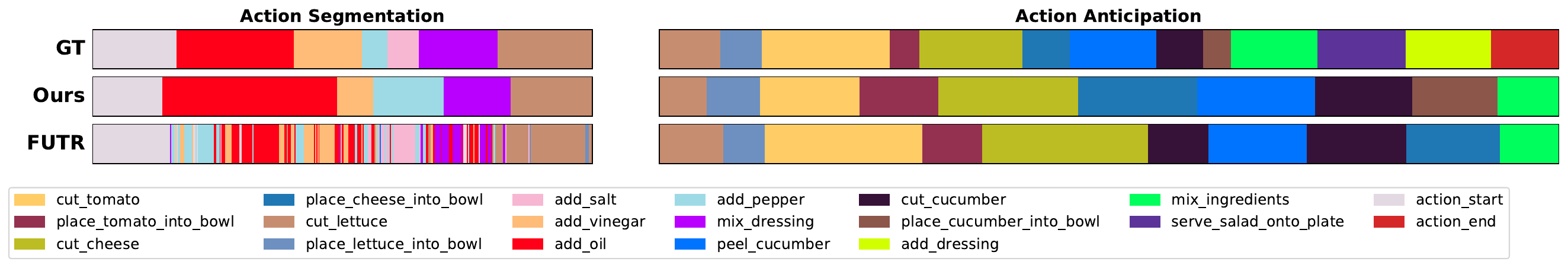}
    \caption{\textbf{50salads.} rgb-10-2; ($\alpha$=0.3, $\beta$=0.5)}
    \label{fig:add_qua_f}
    \end{subfigure}
    \vspace{-1mm}
    \begin{subfigure}[t]{\linewidth}
    \includegraphics[width=\columnwidth]{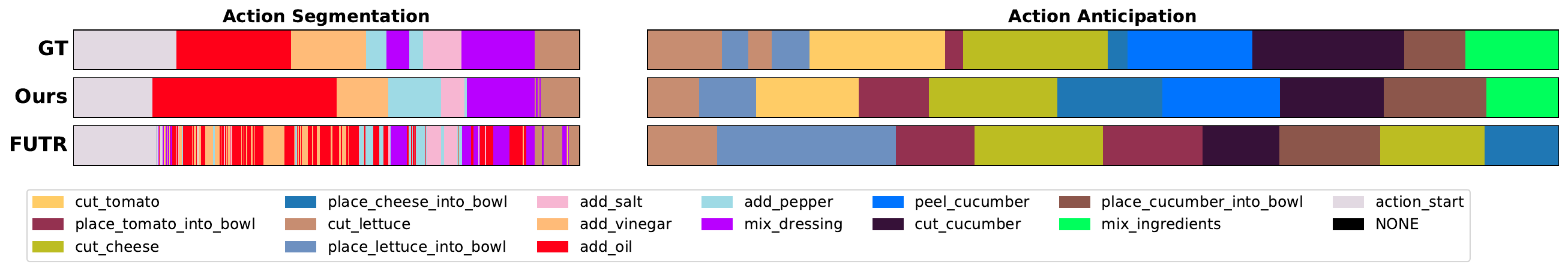}
    \caption{\textbf{50salads.} rgb-10-1; ($\alpha$=0.3, $\beta$=0.5)}
    \label{fig:add_qua_f}
    \end{subfigure}
    \vspace{-2mm}

\caption{\textbf{Additional qualitative results.}  We display the ground-truth, our method (TCCA) and FUTR \cite{FUTR} after using their official checkpoints. The left side of the diagram displays action segmentation from the observation, while the right side shows action anticipation after decoding the action and duration into a frame-wise sequence. Results from different datasets, activities and observation horizon are displayed for comparison.}
\vspace{-4mm}
 \label{fig:add_qualitative}
\end{figure*}

We included additional qualitative results to demonstrate the effectiveness of our approach in a wider range of scenarios, as compared to the FUTR model, as shown in \cref{fig:add_qualitative}. Our method produces more coherent anticipation results due to improved and less noisy segmentation, as well as a better understanding of temporal action constraints.

\end{document}